\documentclass[letterpaper]{article} 
\usepackage{aaai2026}  
\usepackage{times}  
\usepackage{helvet}  
\usepackage{courier}  
\usepackage[hyphens]{url}  
\usepackage{graphicx} 
\usepackage{natbib}  
\usepackage{caption} 
\usepackage{algorithm}
\usepackage{booktabs}
\usepackage{multicol, multirow}
\usepackage{amsmath}
\usepackage{xcolor}         
\usepackage{colortbl}
\usepackage{pgfplots}
\usepackage{tikz}
\usepackage{amssymb}
\usepackage{physics}
\usepackage{graphicx}
\usepackage{subfigure}
\usepackage{algpseudocode}
\usepackage{pgfplotstable}
\usepgfplotslibrary{groupplots}
\usepackage{cleveref}
\usepackage{subcaption}

\definecolor{lightred}{RGB}{255, 204, 204}
\definecolor{lightblue}{RGB}{204, 229, 255}
\definecolor{lightgreen}{RGB}{194, 255, 194}
\definecolor{lightyellow}{RGB}{255, 255, 204}
\definecolor{lightgray}{gray}{0.9}
\definecolor{lightorange}{RGB}{255, 230, 153}

\usepackage{newfloat}
\usepackage{listings}
\DeclareCaptionStyle{ruled}{labelfont=normalfont,labelsep=colon,strut=off} 
\floatstyle{ruled}
\newfloat{listing}{tb}{lst}{}
\floatname{listing}{Listing}
\title{Forget Less by Learning from Parents Through Hierarchical Relationships}
\author {
    Arjun Ramesh Kaushik,
    Naresh Kumar Devulapally,
    Vishnu Suresh Lokhande,
     Nalini K. Ratha,
     Venu Govindaraju
}
\affiliations {
    University at Buffalo-SUNY\\
    \{kaushik3, devulapa, vishnulo, nratha, govind\}@buffalo.edu
}

\usepackage{bibentry}

\begin{document}

\maketitle

\begin{abstract}
Custom Diffusion Models (CDMs) offer impressive capabilities for personalization in generative modeling, yet they remain vulnerable to catastrophic forgetting when learning new concepts sequentially. Existing approaches primarily focus on minimizing interference between concepts, often neglecting the potential for positive inter-concept interactions. In this work, we present Forget Less by Learning from Parents (FLLP), a novel framework that introduces a parent-child inter-concept learning mechanism in hyperbolic space to mitigate forgetting. By embedding concept representations within a Lorentzian manifold, naturally suited to modeling tree-like hierarchies, we define parent-child relationships in which previously learned concepts serve as guidance for adapting to new ones. Our method not only preserves prior knowledge but also supports continual integration of new concepts. We validate FLLP on three public datasets and one synthetic benchmark, showing consistent improvements in both robustness and generalization. 
\end{abstract}


\section{Introduction}
Custom diffusion models (CDMs) \cite{cdm1, mix_of_show} have emerged as a powerful tool for personalizing image generation by adapting general-purpose text-to-image models to individual users or specific concepts. These models enable the synthesis of images from user-provided reference images, offering a pathway to generate personalized content. The growing interest in model customization reflects a broader goal of user-centric generative modeling, where models are tailored to represent distinct objects, styles, or identities that are not present in the training data.
Text-conditioning enhances controllability and flexibility in diffusion models~\cite{ldm_text1, rifegan} by guiding these models to generate semantically aligned visual content. Early personalization methods achieved this by fine-tuning a subset of model weights~\cite{yu2024dreamsteererenhancingsourceimage} or by learning new token embeddings~\cite{cdm1} from a handful of user images. While effective, these methods necessitate per-concept optimization, resulting in increased computational overhead and significant time requirements. Recent work addresses these limitations through tuning-free methods~\cite{wang2024multiclasstextualinversionsecretlyyields}, without the need for re-training during inference.
Beyond individual concept customization, recent research has focused on multi-concept personalization, composing multiple novel concepts into a single generation task~\cite{customconcept101, clora, cidm}. Compared to single-concept settings, multi-concept customization entails diverse reference images and complex compositional prompts 
, leading to challenges such as handling previously unseen concepts, reducing interference among concepts, and maintaining prior knowledge across tasks. These challenges align naturally with the goals of continual learning, where a model learns incrementally from a sequence of concept-specific tasks without catastrophic forgetting. Recent mitigation strategies for catastrophic forgetting in CDMs, such as elastic weight consolidation, latent replay layers, and gradient matching techniques, primarily focus on isolating concepts through regularization and memory optimization. While these methods reduce immediate interference, they often fail to harness the latent relationships between learned concepts \cite{guo2025conceptguard, cidm}. 
Unlike current machine learning systems, human learning naturally excels at transferring knowledge across tasks, an ability rooted in our capacity to organize concepts into hierarchical structures. Research shows that we dynamically update them to support both long-term retention and generalization. Further, recent findings reveal that the brain itself is wired to accumulate and refine such hierarchical conceptual knowledge \cite{intro_neuro1}.


Motivated by these insights, we introduce \textbf{Forget Less by Learning from Parents (FLLP)}, the first framework to leverage inter-concept interactions positively, to the best of our knowledge. \textbf{FLLP} explicitly models hierarchical concept relationships to emulate human-like knowledge transfer. Like cognitive hierarchies, our approach projects concept embeddings into hyperbolic space \cite{anderson2006hyperbolic}, a natural fit for encoding hierarchical structures due to its intrinsic tree-like geometry \cite{meru, sala2018representation}. Within this manifold, we enforce hierarchical parent entailment, requiring new concepts to reside within the entailment cones of their predecessors. This constraint mirrors how humans anchor novel skills (e.g., motorcycle control) to foundational schemas (bicycle balance), preserving conceptual dependencies while enabling adaptive reuse \cite{intro_neuro1}.

\textbf{Our work makes three core contributions}: (1) FLLP leverages natural hierarchies formed by hyperbolic embeddings to model inter-concept interactions. (2) We propose a Hierarchical Parent Entailment Loss using the union-find algorithm \cite{union_find} to form a parent-chain that enforces new concepts to lie within the parent entailment cones. (3) We utilize this hierarchical learning to train CDMs continually and achieve SOTA results across three public datasets and one synthetic benchmark. To the best of our knowledge, this is the first contribution towards \textit{leveraging} inter-concept interactions rather than mitigating them for a continual learning problem in CDMs. 
\begin{figure*}[t!] 
    \centering
    \includegraphics[width=1\linewidth]{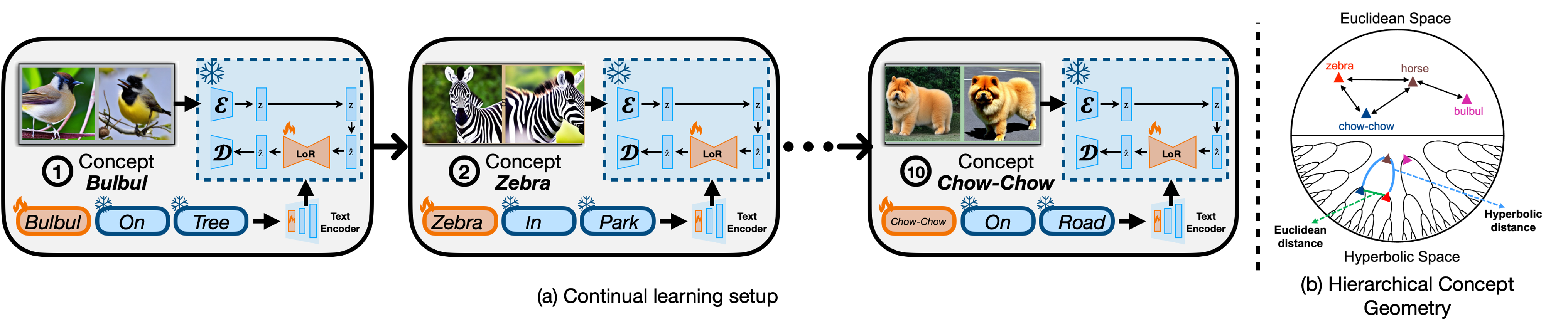}
    \caption{\textbf{Continual Learning for Concept Personalization and Hyperbolic Geometry.} (a) The Continual Learning (CL) paradigm involves training a model on a sequence of tasks or concepts (e.g., Bulbul, Zebra, Chow-chow) over time. A major challenge in this setup is Catastrophic Forgetting, where the model tends to lose knowledge of previously learned concepts as it acquires new ones. (b) In Euclidean space, all concepts coexist without any inherent hierarchy. In contrast, hyperbolic geometry enables hierarchical representation of concepts. Importantly, hyperbolic distances (\textcolor{blue}{blue line}) are fundamentally different from Euclidean distances (\textcolor{teal}{green line}).}
    \label{fig:intro_hb_space}
\end{figure*}

\section{Related Work}
\label{sec:relatedWork}

\noindent\textbf{Incremental Concept Learning.}
Recent class-incremental learning methods focus on mitigating catastrophic forgetting via regularization, distillation, and rehearsal, but are largely developed for discriminative classifiers with fixed label spaces rather than generative models with continuous concept manifolds ~\cite{incLearningSurvey}. The survey in ~\cite{incLearningSurvey} emphasizes that interference between related classes remains a central limitation even under strong regularization or memory mechanisms. In text-to-image diffusion, lifelong and personalized approaches such as L2DM \cite{l2dm}, C-LoRA \cite{clora}, and ConceptGuard \cite{guo2025conceptguard} adapt these ideas to the noise-prediction space by attaching concept-aware adapters and distillation losses to preserve old concepts during continual customization. However, they still treat each concept as an essentially independent unit, offering only coarse control over intra-concept variation and higher-order concept relationships.

\noindent\textbf{Intra-Concept Learning Dynamics.}
Multi-concept customization work shows that concept-specific updates interact in highly non-linear ways. Multi-Concept Customization, Mix-of-Show, and Multi-LoRA all report that composing several learned concepts or LoRA adapters can lead to strong interference and style leakage, especially when concepts are visually similar ~\cite{customconcept101,mix_of_show,multi_lora}. Cross-Attention Bootstrapping further reveals that cross-attention heads in diffusion models specialize to distinct object and attribute factors, and that misaligned heads are a key source of spurious edits and concept entanglement ~\cite{zhang2023crossattention}. Together, these results motivate architectures that explicitly separate and recombine sub-concepts rather than treating each personalized token or adapter as monolithic.

\noindent\textbf{Catastrophic Forgetting in Concept Diffusion.}
Within text-to-image diffusion specifically, forgetting and confusion arise because concept information is distributed across the UNet, cross-attention layers, and token embeddings and is refined across timesteps. L2DM and C-LoRA retain concepts by distilling the denoising trajectory of earlier models and routing gradients through concept-specific low-rank adapters ~\cite{l2dm,clora}, while ConceptGuard disentangles forgetting from confusion and shows that overlapping attention patterns between related concepts are a dominant failure mode ~\cite{guo2025conceptguard}. These systems markedly reduce catastrophic forgetting, but their parameter cost grows roughly linearly with the number of concepts and they still provide only implicit control over inter- and intra-concept interactions.

\noindent\textbf{Hyperbolic Geometry.}
Standard latent diffusion models operate in Euclidean latent spaces ~\cite{ldm}, which are poorly suited for representing deep hierarchies and tree-like concept structure. Hyperbolic vision-language models such demonstrate that negative curvature and exponential volume growth enable compact, hierarchy-aware image-text embeddings with improved semantic organization ~\cite{meru}. HypDiff ~\cite{fu2024hypdiff} proposes a hyperbolic latent diffusion model for graph generation, so as to preserve topological properties such as hierarchy and community structure during generation. However, these methods largely treat hyperbolic geometry as a static representation choice and do not address continual adaptation or concept interference in text-to-image diffusion.

\paragraph{\underline{Limitations of Current Approaches.}}
Across these lines of work, three gaps remain: (1) continual diffusion methods prioritize per-concept preservation but model concepts as independent, limiting intra-concept specialization and compositional recombination ~\cite{l2dm,clora,guo2025conceptguard}; (2) multi-concept customization exposes strong interference between learned edits without providing a principled latent structure for resolving it ~\cite{customconcept101,mix_of_show,multi_lora}; and (3) hyperbolic generative models exploit hierarchical geometry but are not designed for dynamic, concept-aware updates over time ~\cite{meru,liu2023hyperdiffusion}. FLLP targets this intersection by coupling concept-aware continual updates with an explicitly structured latent space that is designed to accommodate evolving concept hierarchies.

\section{Problem Definition}
\label{sec:problem}
We consider the problem setting introduced in CIFC \cite{cidm}, which considers a sequential inflow of concepts. The model learns from an undefined series of text-guided concept customization tasks $T = \{T_c\}_{c=1}^{C}$, where $C$ denotes the total number of tasks. Each task $T_c$ consists of a dataset $T_c = \{(x_c^k, p_c^k, Y_c)\}_{k=1}^{N}$ where $N$ is the number of reference images in the task, $x_c^k$ is an image, $p_c^k$ is a text prompt (e.g., ``photo of a [V$_{cat}$]''), and $Y_c$ represents the concept space. The setting supports diverse customization tasks, including multi-concept generation \cite{multi_lora}, style transfer \cite{zhang2023inversionbasedstyletransferdiffusion}, and image editing \cite{chen2024anydoorzeroshotobjectlevelimage}. There are three constraints in this problem - (1) \textbf{\textit{All concepts are distinct:}} $ Y_c \cap \left(\bigcup_{i=1}^{C-1} Y_i \right) = \emptyset$ indicating that new concepts in task $c$ are distinct from those learned in the other $C-1$ tasks. (2) \textbf{\textit{Concepts are learned in the order of inflow:}} $ T = \{ T_1, T_2, \dots, T_C \} $, where the $c$-th concept $T_c$ is learned with access only to prior concepts $\{T_1, \dots, T_{c-1}\}$ for $c = 1, \dots, C$. This models a unidirectional learning flow, preventing access to future unseen concepts. (3) \textbf{\textit{No-replay constraint:}} No memory storage is allocated to retain training images from past tasks, ensuring purely incremental learning of personalized concepts. This setting allows models to continuously adapt to new personalization tasks while being exposed to catastrophic forgetting of prior concepts.

\section{Understanding Catastrophic Forgetting}

Catastrophic forgetting refers to the tendency of neural networks to lose previously acquired knowledge when trained on new tasks. This phenomenon arises from the stability-plasticity dilemma, where neural networks must balance retaining existing information with adapting to new inputs \cite{vandeven2024continuallearningcatastrophicforgetting}. Catastrophic forgetting occurs when parameter updates for a new task $T_{\text{new}}$ overwrite critical weight configurations optimized for prior tasks $T_{\text{old}}$:
\(P_{\text{old}}(W) \ll P_{\text{new}}(W)\), where $W$ represents network weights, and $P$ denotes task performance. This work advances the central hypothesis that catastrophic forgetting can be fundamentally mitigated through the use of hierarchical knowledge transfer mechanisms.

\subsection{1D Gaussian Setup} 
\label{subsec:1d_gaussian}
To enable a systematic and interpretable analysis of catastrophic forgetting, we construct a synthetic one-dimensional dataset that serves as a controlled surrogate benchmark (see \cref{fig:surg_analysis}). This benchmark comprises five sequential tasks, each defined by a Gaussian distribution with a distinct mean and unit variance, thereby emulating the concept drift characteristic of continual learning scenarios. A shared 1D UNet diffusion model is employed to reconstruct the data distribution associated with each task (see appendix for architectural details). To quantify forgetting, we compute the sum of absolute differences between the generated and true means across tasks. This serves as our primary evaluation metric, referred to as the \textit{forgetting rate}, with lower values indicating improved retention of previously learned concepts.

\subsection{What Causes Catastrophic Forgetting?}

\begin{figure}[!t]
    
\begin{tikzpicture}
\begin{groupplot}[
    group style={group size=5 by 3, horizontal sep=0.1cm, vertical sep=0.5cm},
    width=3cm, height=2.6cm,
    xmin=0, xmax=17.5,
    ymin=0.85, ymax=1.5,
    ylabel={},
    xtick = {3,9,15},
    ytick =\empty,
    axis lines=box,
    tick label style={font=\tiny}, 
    title style={font=\footnotesize}, 
    y label style={at={(axis description cs:0.85,0.5)}, anchor=south, font=\footnotesize},
    xmajorgrids=true,
    ymajorgrids=false,
     legend style={
        font=\tiny,          
        cells={anchor=west}, 
        at={(0.5,-0.15)},    
        anchor=north
    },
]
\pgfplotstableread{testdata2.dat}{\datatable}
\pgfplotsinvokeforeach {0,...,14} {
    \pgfmathsetmacro{\plotnum}{#1}
    
    \ifnum\plotnum=0
        \nextgroupplot[title={Task 1}, title style={font=\footnotesize}, ylabel={\shortstack{Continual\\Learning}}, y label style={font= \scriptsize}]
    \else\ifnum\plotnum=1
        \nextgroupplot[title={Task 2}]
    \else\ifnum\plotnum=2
        \nextgroupplot[title={Task 3}]
    \else\ifnum\plotnum=3
        \nextgroupplot[title={Task 4}]
    \else\ifnum\plotnum=4
        \nextgroupplot[title={Task 5}]
    \else\ifnum\plotnum=5
        \nextgroupplot[ylabel={CIDM}, y label style={font= \scriptsize}]
    \else\ifnum\plotnum=10
        \nextgroupplot[ylabel={\shortstack{FLLP\\(Ours)}}, y label style={font= \scriptsize}]
    \else
        \nextgroupplot
    \fi\fi\fi\fi\fi\fi\fi

    \pgfplotstablegetelem{#1}{meantrue}\of{\datatable}
    \pgfmathsetmacro{\meantrue}{\pgfplotsretval}
    \pgfplotstablegetelem{#1}{stdtrue}\of{\datatable} \pgfmathsetmacro{\stdtrue}{\pgfplotsretval}
    \pgfplotstablegetelem{#1}{meangen}\of{\datatable} \let\meangen\pgfplotsretval
    \pgfplotstablegetelem{#1}{stdgen}\of{\datatable} \let\stdgen\pgfplotsretval
    \draw[->, orange, thick] 
    (axis cs:\meangen,1.4) 
    -- (axis cs:\meantrue,1.4);
    \addplot[domain=\meantrue-\stdtrue:\meantrue+\stdtrue, samples=100, fill=violet!50, draw=none, opacity=0.7]
        {1 + 0.25*sin(deg(pi*(x - (\meantrue - \stdtrue))/(2*\stdtrue)))};
    \addplot[domain=\meantrue-\stdtrue:\meantrue+\stdtrue, samples=2, color=violet!70, thick] {1};
    \addplot[dashed, thick, color=violet] coordinates {(\meantrue,0.75) (\meantrue,1.5)};

    \addplot[domain=\meangen-\stdgen:\meangen+\stdgen, samples=100, fill=orange!60, draw=none, opacity=0.7]
        {1.1 + 0.25*sin(deg(pi*(x - (\meangen - \stdgen))/(2*\stdgen)))};
    \addplot[domain=\meangen-\stdgen:\meangen+\stdgen, samples=2, color=orange!80, thick] {1.1};
    \addplot[dashed, thick, color=orange] coordinates {(\meangen,0.75) (\meangen,1.5)};    
}

\end{groupplot}
\end{tikzpicture}
\caption{\textbf{Quantitative Analysis of 1D Gaussian.} The length of the \textcolor{orange}{orange arrows} reflects the \textit{forgetting rate} for each concept, with shorter arrows indicating better retention. CIDM \cite{cidm} improves upon the baseline by achieving a lower \textit{forgetting rate} of 13.2. FLLP (Ours) further reduces this to 11.4. The visibly shorter \textcolor{orange}{orange arrows} in the bottom row corroborate the improved knowledge retention of FLLP over CIDM. (Best viewed by zooming in) }
\label{fig:surg_analysis}

\end{figure}
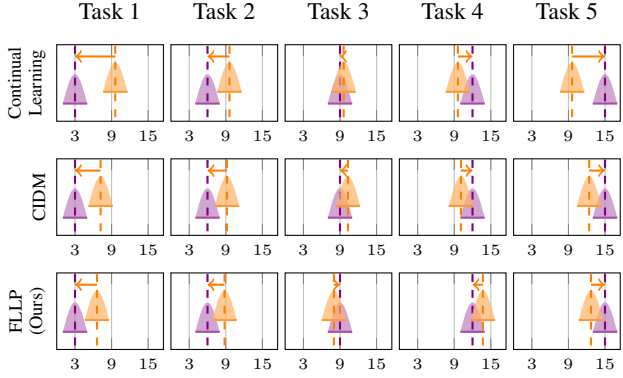

The catastrophic forgetting observed in the 1D Gaussian benchmark \cref{fig:surg_analysis} stems from two interrelated mechanisms: \textit{parameter interference} and \textit{task-specific representation collapse}. When training sequentially on tasks $\mu_i$, gradient updates for new tasks:
$\Delta W \propto \nabla_W \mathcal{L}_{\text{MSE}}(\mu_{\text{new}})$,
perturb the network's shared parameters $W$, overwriting features critical for reconstructing prior distributions $\mu_{\text{old}}$. This interference is amplified by the UNet's frozen architecture, which forces all tasks to compete for representation space through a fixed set of weights.

\textbf{Parameter interference} emerges from fundamental limitations in gradient-based optimization. When learning task $\mu_{\text{new}}$, the model computes weight updates proportional to the gradient of the current task's MSE loss:$\Delta W \propto -\eta \frac{\partial}{\partial W} \|Gen(\mu_{\text{new}}) - \mu_{\text{new}}\|^2$
where $\textit{Gen}$ represents the generated output. These updates create vector fields in parameter space that optimally reduce $\mu_{\text{new}}$'s reconstruction error but catastrophically displace solutions for previous tasks $\mu_{\text{old}}$. In our 1D setup, the frozen UNet architecture compounds this issue; though its core weights remain fixed, the task-specific token embeddings must encode all distributional shifts through a single projection layer. This creates a representational bottleneck where embedding vectors compete to map multiple Gaussian means ($\mu_1,...,\mu_5$) through identical network transformations, leading to destructive interference patterns \cite{lokhande2022equivariance}.

\textbf{Task-specific representation collapse} occurs when sequential training causes embeddings for distinct $\mu_i$ to converge toward a shared subspace. The model's token embeddings $\{E(\mu_i)\}$ gradually align with the dominant gradient directions from new tasks at $t$-th training step: $E(\mu_i)^{t+1} = E(\mu_i)^t - \eta \sum_{j=1}^k \frac{\partial \mathcal{L}_{\text{MSE}}}{\partial E(\mu_j)}$. With no task-specific signals, the model optimizes for a single set of parameters to minimize the aggregate reconstruction error across all tasks. The mean of means emerges as the Nash equilibrium for minimizing the reconstruction error. This is demonstrated in the first row of Fig. \ref{fig:surg_analysis}, where the model converges near the mean-of-means $9.6$, and achieves a \textit{forgetting rate} of $18.6$.

\paragraph{\underline{Analysis:} } Fig. ~\ref{fig:surg_analysis} illustrates the visual outcome of our experiments. Each subplot compares the true and generated distributions for a given task after training on the final task, denoted in blue and red colors, respectively. The matrix layout of subplots is sorted in descending order of \textit{forgetting rate} in different approaches. From top to bottom, we observe a monotonic reduction in error, indicating that the model increasingly preserves information about earlier concepts. The visual alignment of blue (true) and red (generated) bumps in lower rows highlights effective knowledge retention, a desirable trait for continual generative modeling.  Notably, this alignment is most evident in Task 1, Task 2, and Task 3, where the extent of forgetting is also visualized by the length of the arrow indicating the mean shift.

While we have not yet formally introduced our proposed method, FLLP, we offer a brief preview to anchor the reader’s attention. CIDM \cite{cidm} achieves a forgetting rate of $13.2$, whereas our approach, FLLP, lowers this further to $11.4$.  For context, the vanilla continual learning baseline yields a significantly higher forgetting rate of $18.6$. As a reminder, lower forgetting rates indicate better retention of previously learned concepts.


\section{Hyperbolic Trees Enable Optimal Grouping of Concepts}
In this section, we discuss the learning objectives of FLLP to model inter-concept interactions through hyperbolic representations of previously learned concepts and incoming concepts. We begin by introducing hyperbolic geometry and the tools required to implement our methodology. To adapt the $\mathbf{c}$-th customization task $T_c$, we employ LoRA \cite{lora1, lora2} to fine-tune the pretrained denoising UNet, commonly denoted using $\epsilon$. 

\subsection{Background into Hyperbolic Geometry}
\label{subsec:hb_tools}
We adopt the Lorentz model of hyperbolic geometry for developing our framework. In this model, the \( n \)-dimensional hyperbolic space \( \mathbb{H}^n \) is realized as the upper sheet of a two-sheeted hyperboloid embedded in \( \mathbb{R}^{n+1} \), endowed with the Lorentzian inner product (See Sec. \ref{subsec:hyperbolic_bg}). The model’s key advantage lies in its ability to express geodesics, distances, and projections in closed form, which is highly beneficial for optimization in high-dimensional settings. Following the study of special relativity theory \citep{einstein1}, the axis of symmetry of the hyperboloid is referred to as the \textit{time dimension}, while the remaining \( n \) axes form the \textit{space dimensions}. We adopt this terminology throughout, writing vectors in \( \mathbb{R}^{n+1} \) as \( [\mathbf{x}_{\text{space}}, x_{\text{time}}] \). We will now discuss a few tools from hyperbolic geometry to realize the learning objectives for our model.

\paragraph{Geodesics:} The shortest path between two points on a Riemannian manifold is defined as a \textit{geodesic}. In the Lorentz model of hyperbolic space, geodesics are defined as the intersections between the hyperboloid and hyperplanes that pass through the origin of the ambient space $\mathbb{R}^{n+1}$. The Lorentzian distance between two points $x, y \in \mathcal{L}^n$, where $k$ is the constant negative curvature, is given by $
    d_{\mathcal{L}}(x, y) = \frac{1}{\sqrt{k}} \cosh^{-1}(-k \langle x, y \rangle_{\mathcal{L}})$.

\noindent \textbf{Aperture:} The \emph{aperture} defines the half-angle of the entailment cone originating from a point $x$ in hyperbolic space, for a small constant $K$ (typically set to 0.1) is defined as $
\label{eq:aperture}
\texttt{Aper}(x) = \sin^{-1} \left( \frac{2K \sqrt{k}}{\|x_{\text{space}}\|} \right)$.

\noindent \textbf{Exterior Angle:} The \emph{exterior angle} between two Lorentzian embeddings $x$ and $y$ measures how far $y$ lies from the central axis of $x$'s entailment cone. This is defined as the angle between $x$ and $y$ relative to the origin $O$, subtracted from $\pi$, $
\label{eq:ext}
\texttt{Ext}(x, y) = \cos^{-1} \left( \frac{y_{\text{time}} + x_{\text{time}} k \langle x, y \rangle_{\mathcal{L}}}{\|x_{\text{space}}\|  \sqrt{(k \langle x, y \rangle_{\mathcal{L}})^2 - 1}} \right)$

\subsection{Continual Learning in Diffusion Models using Hyperbolic Geometry}

We present a novel learning objective - \textbf{Hyperbolic Parent Entailment Loss} - to model hierarchical relationships between concepts using hyperbolic geometry in Algorithm \ref{alg:entailment_learning} (Appendix). This loss function is governed by three crucial design choices: (1) transferring embeddings from Euclidean space to Lorentz hyperboloid, (2) finding the parent concept(s) for the incoming concept, and (3) designing a training objective to ensure the child concept (incoming concept) is learned through parent guidance (previously learned concepts).

\paragraph{Transferring embeddings to hyperboloid: }
Let an embedding vector be \( \mathbf{v} \in \mathbb{R}^n \). We aim to apply a transformation such that the resulting vector \( \mathbf{x} \) lies on the Lorentz hyperboloid \( \mathcal{L}^n \subset \mathbb{R}^{n+1} \). Let the vector \( \mathbf{v} = [\mathbf{v}, 0] \in \mathbb{R}^{n+1} \). We observe that \( \mathbf{v} \) belongs to the tangent space at the origin \( \mathcal{O} \) of the hyperboloid, since the Lorentzian inner product satisfies:
\(
\langle \mathcal{O}, \mathbf{v} \rangle_\mathcal{L} = 0.
\)
Thus, we parameterize only the space components of the Lorentz model (i.e., \( \mathbf{v} = \mathbf{v}_{\text{space}} \)). This parameterization simplifies the exponential map. Instead of computing the full map in \( \mathbb{R}^{n+1} \), we write the exponential map using only the space components:$ \mathbf{x}_{\text{space}} =  
\frac{\sinh(\sqrt{c} \, \|\mathbf{v}_{\text{space}}\|)}{\sqrt{c} \, \|\mathbf{v}_{\text{space}}\|} \cdot \mathbf{v}_{\text{space}}$. Since the first term is zero, and the Lorentzian norm of \( \mathbf{v} \) simplifies to the Euclidean norm of the space components:
\(
\|\mathbf{v}\|_\mathcal{L}^2 = \langle \mathbf{v}, \mathbf{v} \rangle_\mathcal{L} = \langle \mathbf{v}_{\text{space}}, \mathbf{v}_{\text{space}} \rangle = \|\mathbf{v}_{\text{space}}\|^2,
\)
The corresponding time component \( x_{\text{time}} \) can then be computed from \( \mathbf{x}_{\text{space}} \) as shown in Sec. \ref{subsec:prelim}. The resulting vector \( \mathbf{x} = [x_{\text{time}}, \mathbf{x}_{\text{space}}] \) lies on the Lorentz hyperboloid, eliminating the need for an orthogonal projection step. This parameterization is similar to \cite{meru} and simpler than prior approaches that operate in the full ambient space \( \mathbb{R}^{n+1} \) \citep{pmlr-v97-law19a, le-etal-2019-inferring}.

\begin{figure*}[!h]
    \centering
    \includegraphics[width=1\linewidth]{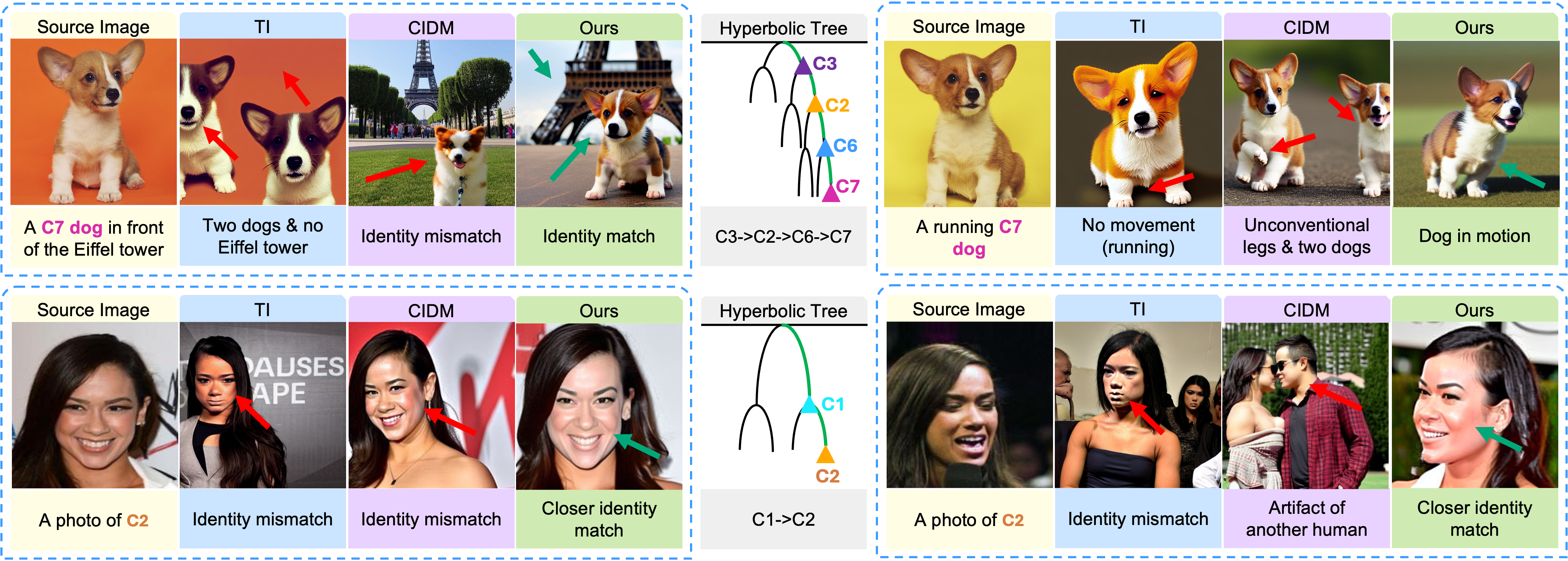}
    \caption{\textbf{Qualitative Analysis.} We compare the generated images in TI \cite{dosovitskiy2021imageworth16x16words}, CIDM \cite{cidm} and FLLP (Ours) on CelebA \cite{celeba} and CIFC \cite{cidm}. The red and green arrows indicate regions of undesirable and desirable qualities, and their reasons are stated below each image. The hyperbolic tree indicates the parent chain of concepts that the model traversed in learning the incoming concept.}
    \label{fig:qual_analysis}
\end{figure*}

\paragraph{Parent Concept Identification: } 
Let the CLIP \cite{clip} image features of the $c$-th concept be $v_c \in \mathbb{R}^{s \times n}$, where $s$ is the number of source images and $n$ is the feature dimensionality. Let $t_c \in \mathbb{R}^{(c-1) \times \text{len}(Y_c)\times n}$ represent the learned token embeddings corresponding to $c-1$ previously known concepts, where $\text{len}(Y_c)$ corresponds to the number of tokens. The image features and learned token embeddings are averaged to obtain $\bar{v_c}\in \mathbb{R}^{1 \times n}$ and $\bar{t_c} \in \mathbb{R}^{(c-1) \times n}$.

We define a recursive search strategy to construct a parent chain for the new concept, denoted by $\mathcal{P_{\text{new}}}$, where each parent will serve as an anchor. The intuitive idea is to iteratively move up the tree from the leaf node to root-level abstractions, as detailed in the Algorithm. \ref{alg:parent-search} (Appendix).

 Let $Z_{\text{img}}$ and $Z_{\text{old}}$ denote the hyperbolic projections of $\bar{v_c}$ and $\bar{t_c}$, respectively. We initialize the current embedding $\mathbf{z}_{\text{curr}} \leftarrow Z_{\text{img}}$ and construct a search pool $\mathcal{S} = Z_{\text{old}} \cup \{Z_{\text{img}}\}$, indexed by $\mathcal{I} = \{0, \dots, c{-}1\} \cup \{-1\}$, where $-1$ refers to the $c$-th concept. At each iteration, we compute the negative Lorentzian geodesic distance between $\mathbf{z}_{\text{curr}}$ and all embeddings in $\mathcal{S}$, selecting the second-closest index (skipping the trivial self-match) as the parent candidate. If the parent index is already visited or corresponds to the new concept itself, we detect a loop and terminate. Otherwise, the selected parent is assigned to $\mathcal{P_{\text{new}}}$, and $\mathbf{z}_{\text{curr}}$ is updated accordingly. This process recursively builds a chain of semantically relevant parent concepts, revealing the new concept’s hierarchical alignment with seen concepts in hyperbolic space. Note that the Lorentzian geodesic distance $d_{\mathcal{L}}$ captures the exponential separation of nodes from the origin.

\paragraph{Storage of Image Attention Maps: }
For each of the $c-1$ previously observed image attention maps, denoted by $I_{\alpha}$ for $\alpha \in \{1, \dots, c-1\}$, we store their timestep-weighted mean representation. This is computed as: $ \bar{I}_{\alpha} =  \sum_{ts=1}^{\text{timesteps}} \frac{1}{\text{ts}} I_{\alpha}^{(ts)}$, where $I_{\alpha}^{(ts)}$ denotes the attention map at timestep $ts$. This aggregation provides a more stable and representative summary of each concept's attention behavior over time. Note that this does not violate the ``no memory" constraint, as the constraint is aimed at restricting the storage of reference images of previous concepts.

\paragraph{Inducing Hierarchical Structure Between Parent-Child Concepts: }
In hyperbolic space, an \textit{entailment cone} $C_u$ at a point $u$ represents the region formed by a hyperbolic cone \cite{pmlr-v80-ganea18a, le-etal-2019-inferring}. These cones narrow as the point $u$ moves farther from the origin. The half-aperture of $C_u$ is defined in Eq.~\ref{eq:aperture}. Let $x$ and $y$ represent the hyperbolic projections (spatial and time components) of the image attention maps corresponding to a child and parent concept, respectively. The associated time components are computed separately, as shown in Sec. \ref{subsec:prelim}. Inspired by prior work on hierarchical embeddings \cite{pmlr-v80-ganea18a}, we utilize entailment cones to enforce a partial ordering between the image attention maps of concepts. In particular, we constrain the child concept $x$ to lie within the entailment cone defined by its parent $y$. To encourage this behavior during training, we introduce a loss term that penalizes violations of this constraint. Specifically, we impose a penalty when the exterior angle between $x$ and $y$ exceeds the aperture of $y$, and $\beta$ is a hyperparameter that controls the flexibility of the constraint:
\begin{equation}
    \mathcal{L}_{\text{entailParent}} = \max\left(0, \texttt{ext}(y, x) - \beta*\texttt{aper}(y)\right)
\end{equation}

\subsection{Training}
Following CIDM \cite{cidm}, we capture semantic patterns that persist in different tasks through direct supervision of LoRA weights. Specifically, a layer-wise common subspace \( \mathbf{W}_*^l \) is introduced to share across tasks. Additionally, a learnable projection matrix \( \mathbf{H}_i^l \) encodes common semantic attributes. Together, we optimize by minimizing the reconstruction loss: $ \mathcal{L}_1 = \sum_{i=1}^{g} \sum_{l=1}^{L} \left\| \Delta \mathbf{W}_i^l - \mathbf{H}_i^l \mathbf{W}_*^l \right\|_F^2$. Finally, we modify the total loss function for the Custom Diffusion model (Refer Sec. \ref{subsec:prelim} in appendix) with \( \mathbf{m}_c = \{\mathbf{m}_c^{l}\}_{l=1}^{L} \) representing layer-wise textual embeddings, and hyperparameter \(\gamma\). Our proposed framework, FLLP, is trained on $\mathcal{L_\text{total}}$ utilizing a dual-step optimization strategy \cite{cidm}. 
{\footnotesize
\begin{equation}
\label{eq:total_loss}
\mathcal{L}_{\text{total}} = \mathbb{E}_{\substack{
z \sim \Phi(x_c^k), m_c \\
\epsilon \sim \mathcal{N}(0, I), t
}} \left[ \| \epsilon - \epsilon_{\theta'_c} (z_t | m_c, t) \|_2^2 + \gamma_1 \mathcal{L}_{\substack{\text{entail} \\ \text{Parent}}} + \gamma_2 \mathcal{L}_1 \right]
\end{equation}}


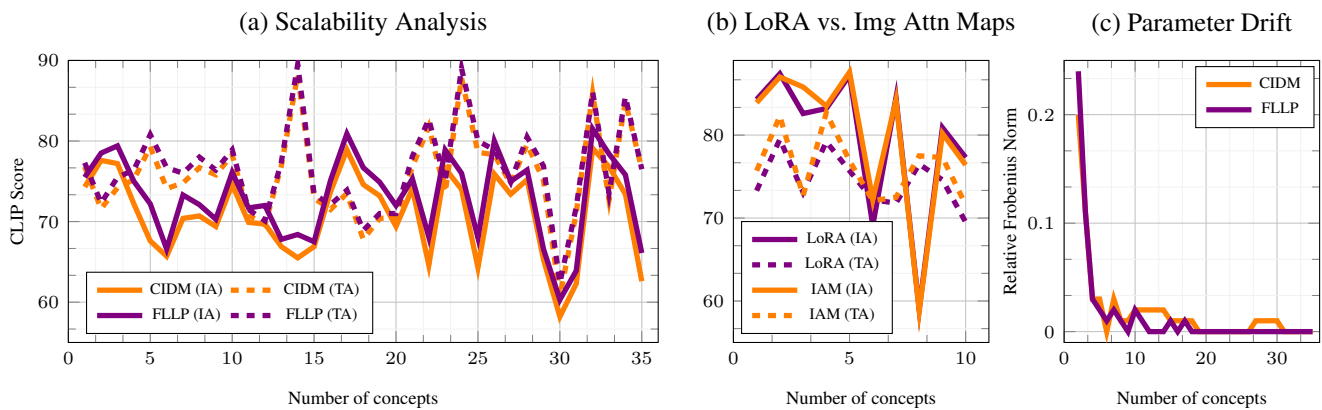
\begin{figure*}[!t]
\centering
\begin{tikzpicture}
\begin{groupplot}[
  group style={
    group size=3 by 1,
    horizontal sep=1cm,
  },
  width=0.28\textwidth,
  height=0.3\textwidth,
  xlabel={Number of concepts},
  xmin=0, xmax=1.1,
  ymin=55, ymax=90,
  grid=both,
  grid style={line width=.1pt, draw=gray!10},
  major grid style={line width=.2pt,draw=gray!50},
  minor tick num=2,
  tick label style={font=\scriptsize},
  label style={font=\scriptsize},
  legend style={font=\tiny},
]

\nextgroupplot[title={(a) Scalability Analysis}, legend columns =2, ylabel={CLIP Score}, xmin=0, xmax=36, width = 0.53\textwidth, y label style={at={(axis description cs:0.07,0.5)}}, legend pos=south west]
\addplot[orange, line width=2pt] table[x=Num, y=IA] {
Num IA
1 74.3
2 77.6
3 77.2
4 72.2
5 67.6
6 65.7
7 70.4
8 70.7
9 69.4
10 74.6
11 69.9
12 69.7
13 66.9
14 65.5
15 66.9
16 73.7
17 78.9
18 74.6
19 73.2
20 69.4
21 74.1
22 64.6
23 76.6
24 74
25 64.3
26 75.9
27 73.4
28 75.2
29 65.3
30 58.2
31 62.3
32 79.1
33 76.8
34 73.4
35 62.6
};
\addlegendentry{CIDM (IA)}

\addplot[dashed, orange, line width=2pt] table[x=Num, y=TA] {
Num TA
1 76.8
2 71.6
3 74.2
4 75.2
5 79.3
6 74.1
7 74.8
8 76.7
9 75.8
10 78.3
11 70.8
12 69.6
13 76.9
14 88.2
15 72.9
16 71.5
17 73.5
18 67.8
19 70.4
20 70.5
21 76.8
22 81.8
23 73.6
24 87.8
25 78.6
26 78.3
27 74.7
28 79.6
29 75.1
30 60.7
31 71.1
32 85.9
33 72.7
34 84.9
35 76.5
};
\addlegendentry{CIDM (TA)}

\addplot[violet, line width=2pt] table[x=Num, y=IA] {
Num IA
1 75.5
2 78.5
3 79.4
4 75
5 72.2
6 66.6
7 73.3
8 72.1
9 70.3
10 76.1
11 71.7
12 72
13 67.8
14 68.4
15 67.5
16 75.1
17 80.9
18 76.7
19 74.9
20 72
21 75.3
22 68.1
23 78.9
24 76
25 68
26 80
27 75
28 76.4
29 66.6
30 60.3
31 63.9
32 81.5
33 78.4
34 75.8
35 66.1
};
\addlegendentry{FLLP (IA)}

\addplot[dashed, violet, line width=2pt] table[x=Num, y=TA] {
Num TA
1 77.3
2 72.3
3 75.5
4 76.4
5 80.8
6 76.7
7 76
8 78
9 76.4
10 78.9
11 71.5
12 70.1
13 77.4
14 89.5
15 73.4
16 72.1
17 73.9
18 68.9
19 71
20 71
21 78.1
22 82.7
23 75
24 89
25 80
26 78.9
27 74.9
28 80.5
29 76.9
30 62.4
31 71.8
32 86.1
33 73.5
34 85.5
35 76.5
};
\addlegendentry{FLLP (TA)}

\nextgroupplot[title={(b) LoRA vs. Img Attn Maps}, ymin = 55, ymax = 89, xmin =0, xmax=11, legend pos = south west]
\addplot[violet, line width=2pt] table[x=Num, y=IA] {
Num IA
1 84.3
2 87.4
3 82.6
4 83.2
5 87.2
6 69.1
7 84.5
8 59.0
9 80.9
10 77.3
};
\addlegendentry{LoRA (IA)}

\addplot[dashed, violet, line width=2pt] table[x=Num, y=TA] {
Num TA
1 73.3
2 79.5
3 73.1
4 79.2
5 75.7
6 72.2
7 71.8
8 76.5
9 74.6
10 69.5
};
\addlegendentry{LoRA (TA)}

\addplot[orange, line width=2pt] table[x=Num, y=IA] {
Num IA
1 83.9
2 87.0
3 85.8
4 83.5
5 87.6
6 72.2
7 84.2
8 58.8
9 80.2
10 76.4
};
\addlegendentry{IAM (IA)}

\addplot[dashed, orange, line width=2pt] table[x=Num, y=TA] {
Num TA
1 75.7
2 82.3
3 72.7
4 82.7
5 76.9
6 72.1
7 72.5
8 77.5
9 77.3
10 71.6
};
\addlegendentry{IAM (TA)}

\nextgroupplot[title={(c) Parameter Drift}, ymax=0.25, ymin=-0.01, xmin=0, xmax = 36, ylabel = {Relative Frobenius Norm}, y label style={at={(axis description cs:0.15,0.5)}}]

\addplot[orange, line width = 2pt] table[x=Num, y=Norm] {
      Num Norm
      2 0.2
      3 0.11
      4 0.03
      5 0.03
      6 0
      7 0.03
      8 0.01
      9 0.01
      10 0.02
      11 0.02
      12 0.02
      13 0.02
      14 0.02
      15 0.01
      16 0.01
      17 0.01
      18 0.01
      19 0
      20 0
      21 0
      22 0
      23 0
      24 0
      25 0
      26 0
      27 0.01
      28 0.01
      29 0.01
      30 0.01
      31 0
      32 0
      33 0
      34 0
      35 0
    };
    \addlegendentry{CIDM}

    \addplot[violet, line width = 2pt] table[x=Num, y=Norm] {
      Num Norm
      2 0.24
      3 0.11
      4 0.03
      5 0.02
      6 0.01
      7 0.02
      8 0.01
      9 0
      10 0.02
      11 0.01
      12 0
      13 0
      14 0
      15 0.01
      16 0
      17 0.01
      18 0
      19 0
      20 0
      21 0
      22 0
      23 0
      24 0
      25 0
      26 0
      27 0
      28 0
      29 0
      30 0
      31 0
      32 0
      33 0
      34 0
      35 0
    };
    \addlegendentry{FLLP}


\end{groupplot}
\end{tikzpicture}
\caption{\textbf{Ablation Studies.} (a) FLLP scales to 35 concepts (limited by CLIP tokenizer) and consistently outperforms CIDM \cite{cidm} on both IA and TA metrics. (b) Hyperbolic Parent Entailment Loss on LoRA weights benefits IA scores, but hurts TA scores. Thereby making Image Attention Maps (IAM) better suited for the loss function. (c) FLLP exhibits 22\% lower parameter drift than CIDM, indicating the utilization of inter-concept interactions positively.}
\label{fig:ablation_main_paper}
\end{figure*}

\section{Experiments}
To rigorously demonstrate the robustness and generalizability of our proposed model, we conduct comprehensive experiments across three diverse public datasets, evaluating performance using CLIP \cite{clip} Image Alignment (IA) and Text Alignment (TA) metrics. Each class is treated as a distinct concept in our continual learning setup. Instead of leveraging all available classes, we intentionally select a subset of ten to simulate the sequential arrival of new concepts, an essential characteristic of realistic continual learning. This controlled design is critical for isolating and faithfully evaluating the effects of catastrophic forgetting under meaningful, yet analytically tractable, conditions.
\noindent\textbf{Remark (Scope of Concept Evaluation):}
In this work, we focus our analysis on a fixed set of 10 concepts across all experiments. This choice is informed by the architectural constraint imposed by the CLIP tokenizer, which allows only 77 learnable tokens \cite{clip}. Since each concept typically consumes 2–3 tokens, the upper bound for feasible training is approximately 35 concepts. While we do present an ablation study extending beyond 10 concepts to examine scalability, our core evaluation remains grounded in 10 concepts, a practical and representative regime for studying catastrophic forgetting. Importantly, our methodology does not inherently limit extension to larger concept sets. Implementation and dataset details are provided in the Appendix.


\subsection{Quantitative and Qualitative Analysis}

\begin{table*}[!ht]
    \centering
    \renewcommand{\arraystretch}{1.5}
    \resizebox{1\textwidth}{!}{
    \begin{tabular}{cc|cc|cc|cc|cc|cc|cc|cc|cc|cc|cc|cc}
        \hline
           \hline
        & \multirow{2}{*}{\textbf{Methods}} & \multicolumn{2}{|c|}{\textbf{C1}} & \multicolumn{2}{|c|}{\textbf{C2}} & \multicolumn{2}{|c|}{\textbf{C3}} & \multicolumn{2}{|c|}{\textbf{C4}} & \multicolumn{2}{|c|}{\textbf{C5}} & \multicolumn{2}{|c|}{\textbf{C6}} & \multicolumn{2}{|c|}{\textbf{C7}} & \multicolumn{2}{|c|}{\textbf{C8}} & \multicolumn{2}{|c|}{\textbf{C9}} & \multicolumn{2}{|c|}{\textbf{C10}} & \multicolumn{2}{|c}{\textbf{Avg.}} \\
        
        \cline{3-24}
        
         & & \textbf{IA} & \textbf{TA} & \textbf{IA} & \textbf{TA} & 
         \textbf{IA} & \textbf{TA} & 
         \textbf{IA} & \textbf{TA} & 
         \textbf{IA} &  \textbf{TA} & 
         \textbf{IA} &  \textbf{TA} & 
         \textbf{IA} &  \textbf{TA} & 
         \textbf{IA} &  \textbf{TA} & 
         \textbf{IA} &  \textbf{TA} & 
         \textbf{IA} &  \textbf{TA} & 
         \textbf{IA} &  \textbf{TA} \\
         
         \hline
          \cellcolor{lightyellow} & \multicolumn{23}{c}{\textbf{Previous frameworks}}\\
         
         \cellcolor{lightyellow}& \textbf{Finetuning} & 
         77.6 &  64.4 & 
         82.2 &  74.6 & 
         79.0 &  69.4 & 
         77.6 &  68.6 & 
         79.6 &  75.0 & 
         62.9 &  70.0 & 
         71.5 &  \underline{76.7} & 
         53.7 &  69.2 & 
         81.4 &  65.4 & 
         72.1 &  67.2 & 
         73.7 &  70.0 \\
         
         \cellcolor{lightyellow} & \textbf{EWC} \cite{ewc} & 
         78.7 &  67.1 & 
         83.8 &  77.5 & 
         80.4 &  72.7 & 
         80.3 &  77.9 & 
         80.7 &  76.7 & 
         64.0 &  \underline{72.3} & 
         76.5 &  74.2 & 
         57.1 &  72.0 & 
         \textbf{84.4} &  66.0 & 
         73.1 &  70.4 & 
         75.9 &  72.7 \\
          
          \cellcolor{lightyellow}& \textbf{LWF} \cite{lwf} & 
         80.4 &  70.8 & 
         79.7 &  75.2 & 
         80.9 &  71.0 &  
         77.4 &  77.4 & 
         80.9 &  76.0 & 
         61.8 &  71.7 & 
         73.2 &  76.3 & 
         53.5 &  72.9 & 
         78.1 &  72.5 &
         74.7 &  70.0 &
         74.1 &  73.4 \\
         
         \cellcolor{lightyellow} & \textbf{CLoRA} \cite{clora} & 
         83.2 &  69.4 & 
         83.4 &  78.0 & 
         81.1 &  \underline{74.1} & 
         80.6 &  78.8 & 
         84.9 &  76.4 & 
         66.3 &  69.6 & 
         76.2 &  \underline{76.7} & 
         58.1 &  73.9 & 
         83.0 &  69.0 & 
         72.1 &  \underline{71.8} & 
         76.9 &  73.6 \\
         
         \cellcolor{lightyellow} & \textbf{L2DM} \cite{l2dm} & 
         78.7 &  68.6 & 
         86.3 &  79.5 & 
         76.6 &  70.1 & 
         80.7 &  73.0 & 
         86.8 &  76.7 & 
         70.8 &  67.7 & 
         70.0 &  75.9 & 
         \textbf{59.3} &  74.1 &
         77.7 &  71.8 & 
         74.1 &  69.4 & 
         76.1 &  72.7\\

         \cellcolor{lightyellow}& \textbf{TI} \cite{dosovitskiy2021imageworth16x16words} & 
          64.9 &   71.8 & 
          56.8 &   59.6 & 
          64.3 &   70.2 & 
          63.1 &   73.1 & 
          61.4  &   66.1 & 
          58.3 &   58.2 & 
          65.7 &   66.1 & 
          57.0 &   57.4 &
          58.7 &   69.7 & 
          64.2 &   68.8 & 
          61.4 &   66.1 \\ 
         \cline{2-24}

         \cellcolor{lightyellow} & \multicolumn{23}{c}{\textbf{Comparison against SOTA framework}}\\
         
         \cellcolor{lightyellow}& \textbf{CIDM}\cite{cidm} & 
         83.6 &  75.3 & 
         86.4 &  78.1 & 
         82.9 &  74.0 & 
         80.8 &  81.1 & 
         86.5 &  78.2 & 
         69.5 &  70.1 & 
         73.7 &  74.7 & 
         56.9 &  74.3 & 
         82.4 &  73.5 & 
         75.9 &  70.2 & 
         78.0 &  74.8 \\
         
         \cellcolor{lightyellow}& \textbf{FLLP (Ours)} & 
          \textbf{83.9} &   \underline{75.7} & 
          \textbf{87.0} &   \underline{82.3} & 
          \textbf{85.8} &   72.7 &  
           \textbf{83.5} &  \underline{82.7} & 
          \textbf{87.6} &   \underline{76.9} & 
          \textbf{72.2} &   72.1 & 
          \textbf{84.2} &  72.5 & 
          58.8 &   \underline{77.5} & 
          80.2 &  \underline{77.3} & 
          \textbf{76.4} &   71.6 & 
          \textbf{80.0} &  \underline{76.1} \\
         
        \rowcolor{lightgray}
         \multirow{-11}{*}{\cellcolor{lightyellow}\rotatebox{90}{\textbf{CIFC dataset}}} & \textbf{$\Delta$} & 
          +0.3 & +0.4 & 
          +0.6 & +4.2 & 
          +2.9 & -1.3 & 
          +2.7 & +1.6 & 
          +1.1 & -1.3 &
          +2.7 & +2.0 & 
          +10.5 & -2.2 & 
          +1.9 & +3.2 & 
          -2.2 & +3.8 & 
          +0.5 & +1.4 & 
        +2.0 & +1.3 \\
         \hline
           \hline
         \cellcolor{lightgreen}& \multicolumn{23}{c}{\textbf{Previous frameworks}}\\
         \cellcolor{lightgreen}& \textbf{TI} \cite{dosovitskiy2021imageworth16x16words} & 
          44.3 &   \underline{62.4} & 
          47.2 &   \underline{65.2} & 
          49.6 &   \underline{61.5} & 
          52.5 &   57.4 & 
          50.2 &   58.7 & 
          49.8 &   54.7 & 
          51.0 &   53.7 & 
          51.3 &   56.4 &
          41.7 &   \underline{61.9} & 
          50.0 &   \underline{65.3} & 
          48.8 &   59.7 \\ 
         \cline{2-24}
         \cellcolor{lightgreen}& \multicolumn{23}{c}{\textbf{Comparison against SOTA framework}}\\
         
         \cellcolor{lightgreen}& \textbf{CIDM}\cite{cidm} & 
          74.2&  59.0 & 
           67.6&  60.4 & 
          73.0&  59.4 & 
           75.4&  58.2 & 
          70.1&  58.0 & 
          78.1&  59.0 & 
          81.0&  58.4 & 
          70.0&  58.0 & 
          70.1&  62.1 & 
          73.0&  55.4 & 
          73.3&  58.8 \\

         \cellcolor{lightgreen}& \textbf{FLLP (Ours)} & 
          \textbf{77.4}&  58.9 & 
          \textbf{74.7}&  61.7 & 
          \textbf{73.4}&  60.8 & 
          \textbf{78.7} &  \underline{63.0} & 
          \textbf{79.0}&  \underline{60.0} & 
          \textbf{82.4}&  \underline{60.5} & 
          \textbf{84.7}&  \underline{61.5} & 
          \textbf{75.8}&  \underline{62.2} & 
          \textbf{75.4}&  61.7 & 
          \textbf{75.6}&  57.7 & 
          \textbf{77.7} &  \underline{60.8} \\
         
        \rowcolor{lightgray}\multirow{-6}{*}{\cellcolor{lightgreen}\rotatebox{90}{\textbf{CelebA dataset}}}& 
         \textbf{$\Delta$} &  
            +3.2 & -0.1 & 
            +7.1 & +1.3 & 
            +0.4 & +1.4 & 
            +3.3 & +4.8 & 
            +8.9 & +2.0 &
            +4.3 & +1.5 & 
            +3.7 & +3.1 & 
            +5.8 & +4.2 & 
            +5.3 & -0.4 & 
            +2.6 & +2.3 & 
            +4.4 & +2.0\\
        \hline
        \hline
        \cellcolor{lightorange}& \multicolumn{23}{c}{\textbf{Previous frameworks}}\\
        
         \cellcolor{lightorange}& \textbf{TI} \cite{dosovitskiy2021imageworth16x16words} & 
          62.2 &   73.4 & 
          66.7 &   76.5 & 
          62.0 &   62.4 & 
          64.0 &   67.2 & 
          64.1 &   66.1 & 
          67.3 &   \underline{76.0}  & 
          63.1 &   63.6 & 
          61.8 &   \underline{78.2} &
         65.8  &   76.5  & 
          66.2 &   72.7 & 
          64.4 &   71.3 \\ 
         \cline{2-24}
         \cellcolor{lightorange}& \multicolumn{23}{c}{\textbf{Comparison against SOTA framework}}\\
         \cellcolor{lightorange}& \textbf{CIDM}\cite{cidm} & 
          80.8&  \underline{74.1} & 
           71.4&  77.9 & 
           81.3&  81.7 & 
           83.0&  82.6 & 
          79.2&  83.7 & 
          \textbf{84.5}&  69.6 & 
          \textbf{76.7}&  \underline{79.3} & 
          87.5&  73.7 & 
          82.2&  \underline{79.9} & 
          85.0&  \underline{82.8} & 
          81.2&  78.5 \\

          \cellcolor{lightorange}& \textbf{FLLP (Ours)} & 
          \textbf{82.1}&  73.9 & 
          \textbf{73.4}&  \underline{78.9} & 
          \textbf{82.6}&  \underline{82.5} & 
           \textbf{83.1}&  \underline{84.0} & 
          \textbf{79.7}&  \underline{85.1} & 
          82.2&  69.7 & 
          75.5&  79.2 & 
          \textbf{90.0}&  75.0 & 
          \textbf{86.4}&  79.8 & 
          \textbf{87.5}&  80.9 & 
          \textbf{82.3}&   \underline{79.0}\\
          
          \rowcolor{lightgray}\multirow{-6}{*}{\cellcolor{lightorange}\rotatebox{90}{\textbf{ImageNet dataset}}}& 
         \textbf{$\Delta$} &  
            +1.3& -0.2 & 
            +2.0 &+1.0 & 
            +1.3 & +0.8 & 
            +0.1& +1.4 & 
            +0.5& +1.4 &
            -2.3&  +0.1& 
            -1.2& -0.1 & 
             +2.5& +1.3 & 
            +4.4& -0.1 & 
            +2.5& -1.9 & 
            +1.1& +0.5 \\
           \hline
           \hline
    \end{tabular}}
    \caption{\textbf{Quantitative Analysis.} We compare our method \textbf{FLLP} against previous frameworks and the SOTA framework CIDM \cite{cidm} across three datasets - CIFC \cite{cidm}, CelebA \cite{celeba} and ImageNet \cite{imagenet}. We observe an improvement in both Image Alignment (IA) and Text Alignment (TA) scores in all datasets. The higher IA score has been denoted in \textbf{bold}, and the higher TA value has been \underline{underlined}.}
    \label{tab:main_exp}
\end{table*}

\textbf{CIFC Dataset: } As shown in Tab. \ref{tab:main_exp}, we achieve superior IA and TA scores compared to CIDM \cite{cidm} in 9 out of 10 concepts. Notably, FLLP achieved $+9.2$ IA scores on the C7 (Dog2) and $+3.2$, being the largest TA gain on the C9 (Cat2). Over the entire dataset, FLLP demonstrates a gain in both IA and TA scores of $+1.2$ and $1.1$, respectively. Note that each concept is trained with a different value of $\beta$ (Threshold parameter). 
\textbf{CelebA Dataset: } On the CelebA dataset \cite{celeba}, FLLP surpasses CIDM with a notable gain of $+4.4$ in Image Alignment (IA) and $+2.0$ in Text Alignment (TA) scores. Among individual concepts, Person 5 (C5) exhibits the highest IA improvement with a margin of $+8.9$, while Person 8 (C8) shows the most significant TA boost, achieving $+4.2$ over CIDM.
\textbf{ImageNet Dataset: } On the third benchmark dataset, FLLP continues to outperform CIDM, achieving gains of $+1.1$ in IA and $+0.5$ in TA scores. Performance improvements are observed consistently across most concepts, with C9 (Wood Rabbit) exhibiting the largest IA gain of $+4.4$. \textit{FLLP consistently outperforms CIDM across CIFC, CelebA, and ImageNet, showing notable gains in both IA and TA metrics.}


\paragraph{Qualitative Analysis: }We provide a qualitative comparison of the synthesized outputs generated by three approaches in Fig. \ref{fig:qual_analysis}: TI \cite{dosovitskiy2021imageworth16x16words}, CIDM \cite{cidm}, and our proposed FLLP framework, spanning CelebA \cite{celeba} (top row) and CIFC \cite{cidm} (bottom row) datasets. Each image is generated using a source image along with its corresponding text prompt. Colored arrows are overlaid to highlight perceptual differences: green arrows denote regions where the output exhibits desirable qualities such as identity preservation or faithful prompt alignment; red arrows highlight deficiencies such as concept drift, artifact introduction, or semantic inconsistency. Commonly, we observe identity mismatches and semantic misalignment across both datasets in previous frameworks that FLLP corrects. Additionally, we visualize the hyperbolic parent-child structure that FLLP leverages during training through a hyperbolic tree. These trees represent the conceptual trajectory taken by the model when incorporating a new concept. \textit{For example, in the case of C7 dog (from the CIFC dataset), FLLP uses a hierarchical parent chain consisting of {Cat1 (C3), Rubber Duck (C2), and Painting (C6)} to guide the learning of the new concept.} In other words, concepts C2, C3, C6 and C7 form a group. Additional examples have been provided in the Appendix. \textit{FLLP enhances output quality by correcting identity and semantic errors through hierarchical hyperbolic guidance.}

\subsection{Ablation Studies}


\noindent \textbf{Why image attention maps instead of LoRA weights?} Direct hyperbolic optimization on LoRA weights forces a simultaneous geometric transformation of both modalities. Image-related weights benefit, while text-related weights suffer \cite{modality_gap}, resulting in a zero-sum game as elucidated in Fig. \ref{fig:ablation_main_paper} (b) and Table \ref{tab:LoRA_vs_imgAttn} (Appendix).

\noindent \textbf{On the effect of $\beta$:} As illustrated in Fig.~\ref{fig:group-thresholds_cifc},\ref{fig:group-thresholds_celeba}, \ref{fig:group-thresholds_inet} (Appendix) across the three datasets, each concept exhibits a distinct optimal threshold value. The threshold modulates the extent of parental influence on child concepts; lower values enforce stronger guidance, while higher values relax this constraint.

\noindent \textbf{Scalability:} To evaluate FLLP beyond ten concepts, we conduct experiments on the CustomConcept101 dataset~\cite{customconcept101}. Due to CLIP’s tokenizer constraint allowing only 77 additional tokens~\cite{clip}, we are limited to training on 35 concepts (considering 2–3 tokens per concept). As shown in Fig. \ref{fig:ablation_main_paper}(a) and Table \ref{tab:scalability} (Appendix), FLLP outperforms CIDM by +2.1 on average IA scores and +1.0 on average TA scores.

\noindent \textbf{Comparing parameter drifts:} We compare the evolution of LoRA weights in FLLP (Ours) and CIDM~\cite{cidm} using the Frobenius norm averaged across all layers on 35 concepts from the CustomConcept101~\cite{customconcept101} dataset. As in Fig. \ref{fig:ablation_main_paper}(c), FLLP exhibits 22\% lower parameter drift, highlighting positive inter-concept interactions.

\section{Conclusion}
In this work, we introduced FLLP, a novel framework that leverages hyperbolic geometry and parent-child relational priors to alleviate catastrophic forgetting in CDMs. Unlike existing approaches that reduce inter-concept interactions, FLLP exploits structured inter-concept dependencies through a Lorentzian manifold, enabling positive transfer and robust continual adaptation. Across diverse benchmarks, our empirical results demonstrate that FLLP reliably enhances prior-knowledge retention, strengthens generalization, and scales effectively to new concepts.

\bibliography{aaai2026}


\newpage
\appendix
\section{Technical Appendices and Supplementary Material}
\subsection{Inference}
Similar to CIDM, our inference strategy relies on Elastic Weight Aggregation (EWA). The learned LoRA weights are aggregated based on the semantic relationship between the stored concept token embedding ($\hat{c}$) and the current task's prompt embedding ($\hat{e}$) to reduce forgetting. The aggregated low-rank weight \( \Delta \mathbf{\hat{W}}^l \) in the \( l \)-th transformer layer is formulated as: $\Delta \mathbf{\hat{W}}^l = \sum_{i=1}^{c} \Delta \mathbf{W}_i^l \cdot \psi(\mathbf{\mathcal{M}})_i$, and
$\mathbf{\mathcal{M}} = \max (\mathbf{\hat{c}}^l \cdot (\mathbf{\hat{e}}^l)^\top) \quad $, where \( \max(\cdot) \) operates along the row axis, and $ \psi(\mathbf{\mathcal{M}})$ denotes the frobenius-norm of $\mathcal{M}$. After aggregating the learned low-rank weights, we obtain the updated denoising UNet \( \epsilon_{\theta'_{\ast}}(\cdot) \) for inference, where \( \theta'_{\ast} = \theta_0 + \Delta \theta_{\ast} \) and \( \Delta \theta_{\ast} = \{\Delta \mathbf{\hat{W}}^l\}_{l=1}^{L} \). This formulation effectively consolidates the essential characteristics of all personalized concepts.

\subsection{Threshold Affects the Hierarchical Concept Tree}

In this section, we examine the influence of the threshold parameter \( \beta \) on the performance of our framework. As illustrated in Figures~\ref{fig:group-thresholds_cifc}, \ref{fig:group-thresholds_celeba}, and \ref{fig:group-thresholds_inet} across the three datasets, each concept exhibits a distinct optimal threshold value. The threshold modulates the extent of parental influence exerted on child concepts within the hierarchy: lower values of \( \beta \) enforce stronger guidance, while higher values relax this constraint. Although the loss function is applied exclusively to the image attention maps, the concept token embeddings are indirectly shaped by the threshold, consequently altering the structure of the induced concept tree. This behavior is further substantiated in Table~\ref{tab:main_ablation}, which demonstrates that varying the threshold results in different parent-child relationships within the concept hierarchy. Thresholding effectively filters which parent embeddings influence the child, allowing us to balance between inheritance and autonomy. This dynamic control is crucial in continual learning, where over-regularization can harm
generalization, and under-regularization may lead to concept drift.

\begin{table*}[!ht]
    \centering
    \renewcommand{\arraystretch}{1.8}
    \resizebox{1\textwidth}{!}{
    \begin{tabular}{c|c|c|c|c|c|c|c|c|c|c|c}
        \hline
        \hline
        \textbf{Dataset} & \textbf{t} & \textbf{C1} & \textbf{C2} & \textbf{C3} & \textbf{C4} & \textbf{C5} & \textbf{C6} & \textbf{C7} & \textbf{C8} & \textbf{C9} & \textbf{C10}\\
         \hline
          
         \cellcolor{lightyellow} & 0.1 
         & \{\} 
         & \{C1\} 
         & \{C2, C1\} 
         & \{C3, C2\} 
         & \{C3, C2\} 
         & \underline{\textbf{\{C3, C2\}}} & 
         \{C6, C2, C3\} & \underline{\textbf{\{C6, C2, C3\}}} & \underline{\textbf{\{C6, C2, C3\}}} & \{C6, C2, C3\} \\
         
         \cellcolor{lightyellow} &  0.3 & \{\} & \{C1\} & \{C2, C1\} & \{C3, C2\} & \{C3, C2\} & \{C3, C2\} & \{C3, C2\} & \{C3, C2\} & \{C3, C2\} & \{C3, C2\} \\
         
         \cellcolor{lightyellow} & 0.5 & \{\} & \{C1\} & \{C2, C1\} & \{C3, C2\} & \{C3, C2\} & \{C3, C2\} & \{C6, C2, C3\} & \{C3, C2\} & \{C3, C2\} & \underline{\textbf{\{C3, C2\}}} \\
         
         \cellcolor{lightyellow} &  0.7 & \{\} & \{C1\} & \{C2, C1\} & \{C3, C2\} & \underline{\textbf{\{C3, C2\}}} & \{C3, C2\} & \underline{\textbf{\{C6, C2, C3\}}} & \{C6, C2, C3\} & \{C6, C2, C3\} & \{C6, C2, C3\} \\
         
         \cellcolor{lightyellow} & 0.9 & \{\} & \{C1\} & \{C2, C1\} & \{C3, C2\} & \{C3, C2\} & \{C3, C2\} & \{C3, C2, C6\} & \{C3, C2, C6\} & \{C3, C2, C6\} & \{C3, C2, C6\} \\
         
         \multirow{-6}{*}{\cellcolor{lightyellow} \textbf{\rotatebox{90}{\small CIFC \cite{cidm}}}} &  1.0 & \{\} & \underline{\textbf{\{C1\}}} & \underline{\textbf{\{C2, C1\}}} & \underline{\textbf{\{C3, C2\}}} & \{C3, C2\} & \{C3, C2\} & \{C3, C2, C6\} & \{C3, C2, C6\} & \{C3, C2, C6\} & \underline{\textbf{\{C3, C2, C6\}}} \\
         
         \hline
         \cellcolor{lightgreen} & 0.1 & \{\} & \{C1\} & \underline{\textbf{\{C1, C2\}}} & \{C1, C2\} & \{C1, C2\} & \{C1, C5\} & \{C1, C5\} & \{C1, C5\} & \{C1, C5\} & \underline{\textbf{\{C1, C5\}}} \\
         
         \cellcolor{lightgreen} &  0.3 & \{\} & \{C1\} & \{C2, C1\} & \underline{\textbf{\{C2, C1, C3\}}} & \{C2, C1, C4\} & \underline{\textbf{\{C2, C5, C1, C4\}}} & \{C2, C6, C5\} & \{C2, C6, C5\} & \underline{\textbf{\{C2, C6, C5\}}} & \{C2, C6, C5\} \\
         
         \cellcolor{lightgreen} & 0.5 & \{\} & \underline{\textbf{\{C1\}}} & \{C1, C2\} & \{C1, C3\} & \{C1, C3\} & \{C1, C3\} & \{C1, C6, C2\} & \{C1, C6, C2\} & \{C5, C2, C6\} & \{C1, C6, C2\} \\
         
         \cellcolor{lightgreen} &  0.7 & \{\} & \{C1\} & \{C1, C2\} & \{C1, C3\} & \{C1, C3\} & \{C1, C5\} & \{C1, C6, C2\} & \{C1, C6, C2\} & \{C1, C6, C2\} & \{C1, C6, C2\} \\
         
         \cellcolor{lightgreen} & 0.9 & \{\} & \{C1\} & \{C1, C2\} & \{C1, C2\} & \{C1, C2\} & \underline{\textbf{\{C1, C2\}}} & \{C1, C6, C2\} & \underline{\textbf{\{C1, C6, C2\}}} & \{C1, C6, C2\} & \{C1, C6, C2\} \\
         \multirow{-6}{*}{\cellcolor{lightgreen}\rotatebox{90}{\textbf{\small CelebA \cite{celeba}}}} &  1.0 & \{\} & \{C1\} & \{C1, C2\} & \{C1, C3\} & \underline{\textbf{\{C1, C3\}}} & \{C1, C3\} & \{C1, C6, C3\} & \{C1, C6, C3\} & \{C1, C6, C3\} & \{C1, C6, C3\} \\
            \hline
         \cellcolor{lightorange} & 0.1 & \{\} & \underline{\textbf{\{C1\}}} & \{C1, C2\} & \{C2, C3\} & \{C1, C2, C3\} & \underline{\textbf{\{C2, C3\}}} & \{C2, C3\} & \{C2, C3\} & \{C2, C3\} & \underline{\textbf{\{C1, C2, C3\}}} \\
         
         \cellcolor{lightorange}&  0.3 & \{\} & \{C1\} & \{C1, C2\} & \{C1, C2, C3\} & \underline{\textbf{\{C1, C2, C3\}}} & \{C5, C3, C2\} & \{C5, C3, C2\} & \{C5, C3, C2\} & \{C5, C3, C2\} & \{C5, C3, C2\} \\
         
         \cellcolor{lightorange} & 0.5 & \{\} & \{C1\} & \underline{\textbf{\{C1, C2\}}} & \{C1, C2, C3\} & \{C1, C2, C3\} & \{C5, C3, C2\} & \underline{\textbf{\{C5, C3, C2\}}} & \{C5, C3, C2\} & \{C5, C3, C2\} & \{C1, C2, C3\} \\
         
         \cellcolor{lightorange}&  0.7 & \{\} & \{C1\} & \{C2, C1\} & \{C2, C3\} & \{C4, C2, C3\} & \{C4, C2, C3\} & \{C4, C2, C3\} & \{C5, C3, C2\} & \{C5, C3, C2\} & \{C2, C3\} \\
         
         \cellcolor{lightorange}& 0.9 & \{\} & \{C1\} & \{C1, C2\} & \underline{\textbf{\{C3, C2\}}} & \{C3, C2\} & \{C3, C2\} & \{C5, C3, C2\} & \underline{\textbf{\{C5, C3, C2\}}} & \{C3, C2\} & \{C2, C3\} \\
         \multirow{-6}{*}{\cellcolor{lightorange}\rotatebox{90}{\textbf{\small ImageNet \cite{imagenet}}}}&  1.0 & \{\} & \{C1\} & \{C2, C1\} & \{C2, C1\} & \{C2, C3\} & \{C2, C3\} & \{C2, C3\} & \{C2, C3\} & \underline{\textbf{\{C2, C3\}}} & \{C2, C3\} \\
         \bottomrule
         \bottomrule
    \end{tabular}}
    \caption{\textbf{Parent chains formed by each concept at different thresholds (t).} Training the image attention maps at different threshold values also affects the token embeddings. Thereby, influencing the formation of the concept-tree. The parents have been ordered from nearest to farthest hyperbolic distance. The optimal parent chain for each concept has been denoted by \underline{\textbf{bold-underline}}.}
    \label{tab:num_of_parents}
\end{table*}

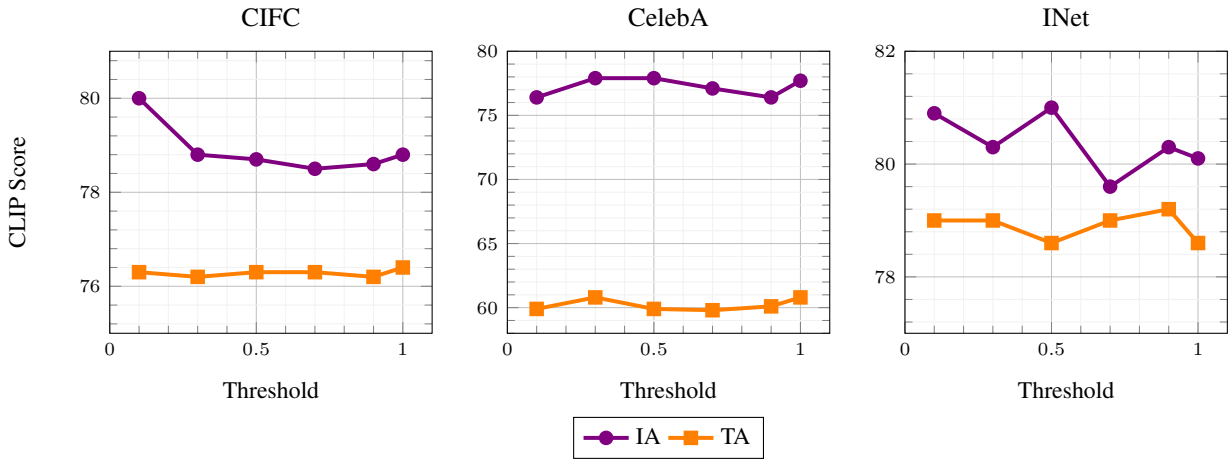
\begin{figure*}[!h]
\centering
\begin{tikzpicture}
\begin{groupplot}[
  group style={
    group size=3 by 1,
    horizontal sep=1cm,
  },
  width=0.33\textwidth,
  height=0.3\textwidth,
  xlabel={Threshold},
  xmin=0, xmax=1.1,
  ymin=70, ymax=81,
  grid=both,
  grid style={line width=.1pt, draw=gray!10},
  major grid style={line width=.2pt,draw=gray!50},
  minor tick num=4,
  tick label style={font=\scriptsize},
  label style={font=\small},
  legend style={font=\footnotesize},
]

\nextgroupplot[title={CIFC}, ymin =75, ymax =81,legend to name=zelda, legend columns =2,
  ylabel={CLIP Score}]
\addplot[mark=*, violet, line width=1.5pt] table[x=Threshold, y=IA] {
Threshold IA
0.1 80.0
0.3 78.8
0.5 78.7
0.7 78.5
0.9 78.6
1 78.8
};
\addlegendentry{IA}
\addplot[mark=square*, orange, line width=1.5pt] table[x=Threshold, y=TA] {
Threshold TA
0.1 76.3
0.3 76.2
0.5 76.3
0.7 76.3
0.9 76.2
1 76.4
};
\addlegendentry{TA}

\nextgroupplot[title={CelebA}, ymax=80, ymin=58]
\addplot[mark=*, violet, line width=1.5pt] table[x=Threshold, y=IA] {
Threshold IA
0.1 76.4
0.3 77.9
0.5 77.9
0.7 77.1
0.9 76.4
1 77.7
};
\addplot[mark=square*, orange, line width=1.5pt] table[x=Threshold, y=TA] {
Threshold TA
0.1 59.9
0.3 60.8
0.5 59.9
0.7 59.8
0.9 60.1
1 60.8
};

\nextgroupplot[title={INet}, ymin = 77, ymax = 82]
\addplot[mark=*, violet, line width=1.5pt] table[x=Threshold, y=IA] {
Threshold IA
0.1 80.9
0.3 80.3
0.5 81.0
0.7 79.6
0.9 80.3
1 80.1
};
\addplot[mark=square*, orange, line width=1.5pt] table[x=Threshold, y=TA] {
Threshold TA
0.1 79.0
0.3 79.0
0.5 78.6
0.7 79.0
0.9 79.2
1 78.6
};

\end{groupplot}
\path (group c1r1.south west) -- (group c3r1.south east) coordinate[midway] (centerbelow);
\node[anchor=north] at (centerbelow |- current bounding box.south) {\pgfplotslegendfromname{zelda}};
\end{tikzpicture}
\caption{\textbf{Average CLIP scores of all ten concepts.} Effect of threshold parameter on the average CLIP Score across the three datasets.}
\label{fig:threshold}
\end{figure*}

\pgfplotsset{compat=1.15}

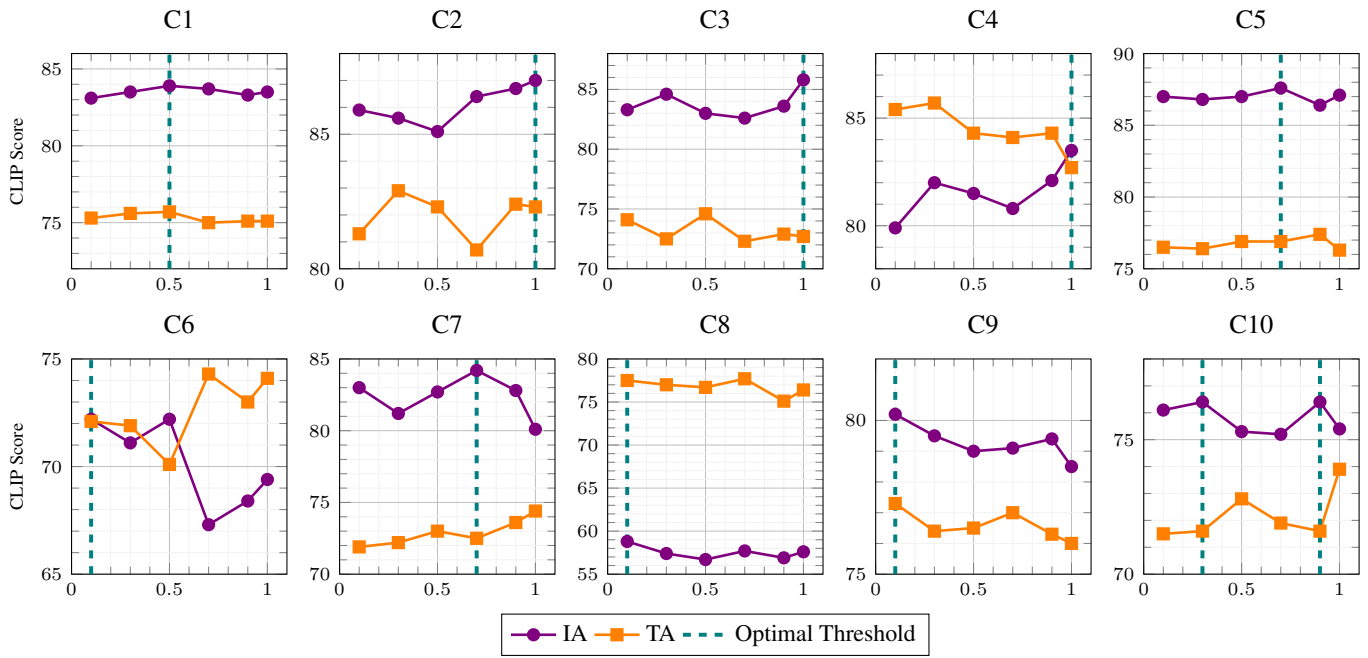
\begin{figure*}[!h]
\centering
\begin{tikzpicture}
\begin{groupplot}[
  group style={
    group size=5 by 2,
    horizontal sep=0.7cm,
    vertical sep=1.2cm,
  },
  width=0.25\textwidth,
  height=0.25\textwidth,
  xmin=0, xmax=1.1,
  ymin=55, ymax=90,
  ytick distance=5,
  label style={font=\scriptsize},
  tick label style={font=\scriptsize},
  grid=both,
  grid style={line width=.1pt, draw=gray!10},
  major grid style={line width=.2pt,draw=gray!50},
  minor tick num=4,
  legend style={font=\small},
]

\nextgroupplot[title=C1, ylabel={CLIP Score}, ymin= 72, ymax=86, legend to name=zelda, legend columns =3]
\addplot[mark=*, violet, line width=1pt] table[x=Threshold, y=IA] {data_cifc1.dat};
\addlegendentry{IA}
\addplot[mark=square*, orange, line width=1pt] table[x=Threshold, y=TA] {data_cifc1.dat};
\addlegendentry{TA}
\addplot[
  dashed,
  thick,
  teal,
  line width = 1.5pt
] coordinates {(0.5, 72) (0.5, 86)};
\addlegendentry{Optimal Threshold}

\nextgroupplot[title=C2, ymin = 80, ymax = 88]
\addplot[mark=*, violet, line width=1pt] table[x=Threshold, y=IA] {data_cifc2.dat};
\addplot[mark=square*, orange, line width=1pt] table[x=Threshold, y=TA] {data_cifc2.dat};
\addplot[
  dashed,
  thick,
  teal,
  line width = 1.5pt
] coordinates {(1, 80) (1, 88)};

\nextgroupplot[title=C3, ymin = 70, ymax=88]
\addplot[mark=*, violet, line width=1pt] table[x=Threshold, y=IA] {data_cifc3.dat};
\addplot[mark=square*, orange, line width=1pt] table[x=Threshold, y=TA] {data_cifc3.dat};
\addplot[
  dashed,
  thick,
  teal,
  line width = 1.5pt
] coordinates {(1, 70) (1, 88)};

\nextgroupplot[title=C4, ymin=78, ymax=88]
\addplot[mark=*, violet, line width=1pt] table[x=Threshold, y=IA] {data_cifc4.dat};
\addplot[mark=square*, orange, line width=1pt] table[x=Threshold, y=TA] {data_cifc4.dat};
\addplot[
  dashed,
  thick,
  teal,
  line width = 1.5pt
] coordinates {(1, 78) (1, 88)};

\nextgroupplot[title=C5, ymin=75]
\addplot[mark=*, violet, line width=1pt] table[x=Threshold, y=IA] {data_cifc5.dat};
\addplot[mark=square*, orange, line width=1pt] table[x=Threshold, y=TA] {data_cifc5.dat};
\addplot[
  dashed,
  thick,
  teal,
  line width = 1.5pt
] coordinates {(0.7, 75) (0.7, 90)};

\nextgroupplot[title=C6, ylabel={CLIP Score}, ymin=65, ymax = 75]
\addplot[mark=*, violet, line width=1pt] table[x=Threshold, y=IA] {data_cifc6.dat};
\addplot[mark=square*, orange, line width=1pt] table[x=Threshold, y=TA] {data_cifc6.dat};
\addplot[
  dashed,
  thick,
  teal,
  line width = 1.5pt
] coordinates {(0.1, 65) (0.1, 75)};

\nextgroupplot[title=C7, ymin=70, ymax=85]
\addplot[mark=*, violet, line width=1pt] table[x=Threshold, y=IA] {data_cifc7.dat};
\addplot[mark=square*, orange, line width=1pt] table[x=Threshold, y=TA] {data_cifc7.dat};
\addplot[
  dashed,
  thick,
  teal,
  line width = 1.5pt
] coordinates {(0.7, 70) (0.7, 85)};

\nextgroupplot[title=C8, ymin=55, ymax=80]
\addplot[mark=*, violet, line width=1pt] table[x=Threshold, y=IA] {data_cifc8.dat};
\addplot[mark=square*, orange, line width=1pt] table[x=Threshold, y=TA] {data_cifc8.dat};
\addplot[
  dashed,
  thick,
  teal,
  line width = 1.5pt
] coordinates {(0.1, 55) (0.1, 80)};

\nextgroupplot[title=C9, ymin=75, ymax=82]
\addplot[mark=*, violet, line width=1pt] table[x=Threshold, y=IA] {data_cifc9.dat};
\addplot[mark=square*, orange, line width=1pt] table[x=Threshold, y=TA] {data_cifc9.dat};
\addplot[
  dashed,
  thick,
  teal,
  line width = 1.5pt
] coordinates {(0.1, 75) (0.1, 82)};

\nextgroupplot[title=C10, ymin=70, ymax=78]
\addplot[mark=*, violet, line width=1pt] table[x=Threshold, y=IA] {data_cifc10.dat};
\addplot[mark=square*, orange, line width=1pt] table[x=Threshold, y=TA] {data_cifc10.dat};
\addplot[
  dashed,
  thick,
  teal,
  line width = 1.5pt
] coordinates {(0.3, 70) (0.3, 78)};
\addplot[
  dashed,
  thick,
  teal,
  line width = 1.5pt
] coordinates {(0.9, 70) (0.9, 78)};
\end{groupplot}

\path (group c1r2.south west) -- (group c5r2.south east) coordinate[midway] (centerbelow);
\node[anchor=north] at (centerbelow |- current bounding box.south) {\pgfplotslegendfromname{zelda}};
\end{tikzpicture}
\caption{\textbf{Effect of threshold variation on the CIFC dataset.} In this plot, we analyse the variation of CLIP scores (\textcolor{violet}{\textbf{IA}} and \textcolor{orange}{\textbf{TA}}) for different threshold values. We observe that each concept has a specific threshold value that exhibits maximum gains in scores.}
\label{fig:group-thresholds_cifc}
\end{figure*}

\pgfplotsset{compat=1.15}

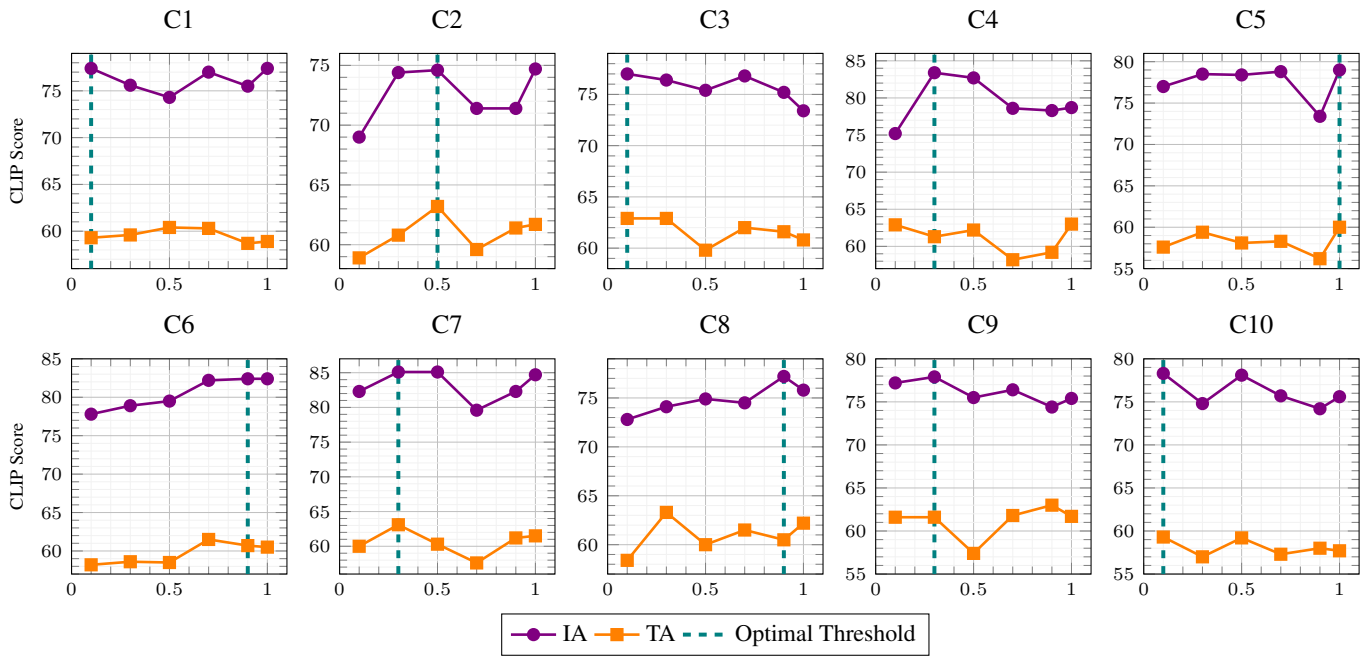
\begin{figure*}[!h]
\centering
\begin{tikzpicture}
\begin{groupplot}[
  group style={
    group size=5 by 2,
    horizontal sep=0.7cm,
    vertical sep=1.2cm,
  },
  width=0.25\textwidth,
  height=0.25\textwidth,
  xmin=0, xmax=1.1,
  ymin=55, ymax=90,
  label style={font=\scriptsize},
  tick label style={font=\scriptsize},
  grid=both,
  ytick distance=5,
  grid style={line width=.1pt, draw=gray!10},
  major grid style={line width=.2pt,draw=gray!50},
  minor tick num=4,
  legend style={font=\small},
]

\nextgroupplot[title=C1, ylabel={CLIP Score}, ymin = 56, ymax = 79, legend to name=zelda, legend columns =3]
\addplot[mark=*, violet, line width=1pt] table[x=Threshold, y=IA] {data_celeba1.dat};
\addlegendentry{IA}
\addplot[mark=square*, orange, line width=1pt] table[x=Threshold, y=TA] {data_celeba1.dat};
\addlegendentry{TA}
\addplot[
  dashed,
  thick,
  teal,
  line width = 1.5pt
] coordinates {(0.1, 56) (0.1, 79)};
\addlegendentry{Optimal Threshold}

\nextgroupplot[title=C2, ymin=58, ymax=76]
\addplot[mark=*, violet, line width=1pt] table[x=Threshold, y=IA] {data_celeba2.dat};
\addplot[mark=square*, orange, line width=1pt] table[x=Threshold, y=TA] {data_celeba2.dat};
\addplot[
  dashed,
  thick,
  teal,
  line width = 1.5pt
] coordinates {(0.5, 58) (0.5, 76)};

\nextgroupplot[title=C3, ymin=58, ymax=79]
\addplot[mark=*, violet, line width=1pt] table[x=Threshold, y=IA] {data_celeba3.dat};
\addplot[mark=square*, orange, line width=1pt] table[x=Threshold, y=TA] {data_celeba3.dat};
\addplot[
  dashed,
  thick,
  teal,
  line width = 1.5pt
] coordinates {(0.1, 58) (0.1, 79)};

\nextgroupplot[title=C4, ymin=57, ymax=86]
\addplot[mark=*, violet, line width=1pt] table[x=Threshold, y=IA] {data_celeba4.dat};
\addplot[mark=square*, orange, line width=1pt] table[x=Threshold, y=TA] {data_celeba4.dat};
\addplot[
  dashed,
  thick,
  teal,
  line width = 1.5pt
] coordinates {(0.3, 57) (0.3, 86)};

\nextgroupplot[title=C5, ymin=55, ymax=81]
\addplot[mark=*, violet, line width=1pt] table[x=Threshold, y=IA] {data_celeba5.dat};
\addplot[mark=square*, orange, line width=1pt] table[x=Threshold, y=TA] {data_celeba5.dat};
\addplot[
  dashed,
  thick,
  teal,
  line width = 1.5pt
] coordinates {(1, 55) (1, 81)};

\nextgroupplot[title=C6, ylabel={CLIP Score}, ymin=57, ymax=85]
\addplot[mark=*, violet, line width=1pt] table[x=Threshold, y=IA] {data_celeba6.dat};
\addplot[mark=square*, orange, line width=1pt] table[x=Threshold, y=TA] {data_celeba6.dat};
\addplot[
  dashed,
  thick,
  teal,
  line width = 1.5pt
] coordinates {(0.9, 57) (0.9, 85)};

\nextgroupplot[title=C7, ymin=56, ymax=87]
\addplot[mark=*, violet, line width=1pt] table[x=Threshold, y=IA] {data_celeba7.dat};
\addplot[mark=square*, orange, line width=1pt] table[x=Threshold, y=TA] {data_celeba7.dat};
\addplot[
  dashed,
  thick,
  teal,
  line width = 1.5pt
] coordinates {(0.3, 56) (0.3, 87)};

\nextgroupplot[title=C8, ymin=57, ymax=79]
\addplot[mark=*, violet, line width=1pt] table[x=Threshold, y=IA] {data_celeba8.dat};
\addplot[mark=square*, orange, line width=1pt] table[x=Threshold, y=TA] {data_celeba8.dat};
\addplot[
  dashed,
  thick,
  teal,
  line width = 1.5pt
] coordinates {(0.9, 57) (0.9, 79)};

\nextgroupplot[title=C9, ymax=80]
\addplot[mark=*, violet, line width=1pt] table[x=Threshold, y=IA] {data_celeba9.dat};
\addplot[mark=square*, orange, line width=1pt] table[x=Threshold, y=TA] {data_celeba9.dat};
\addplot[
  dashed,
  thick,
  teal,
  line width = 1.5pt
] coordinates {(0.3, 55) (0.3, 80)};

\nextgroupplot[title=C10, ymax=80]
\addplot[mark=*, violet, line width=1pt] table[x=Threshold, y=IA] {data_celeba10.dat};
\addplot[mark=square*, orange, line width=1pt] table[x=Threshold, y=TA] {data_celeba10.dat};
\addplot[
  dashed,
  thick,
  teal,
  line width = 1.5pt
] coordinates {(0.1, 55) (0.1, 80)};

\end{groupplot}
\path (group c1r2.south west) -- (group c5r2.south east) coordinate[midway] (centerbelow);
\node[anchor=north] at (centerbelow |- current bounding box.south) {\pgfplotslegendfromname{zelda}};
\end{tikzpicture}
\caption{\textbf{Effect of threshold variation on the CelebA dataset.} In this plot, we analyse the variation of CLIP scores (\textcolor{violet}{\textbf{IA}} and \textcolor{orange}{\textbf{TA}}) for different threshold values. We observe that each concept has a specific threshold value that exhibits maximum gains in scores.}
\label{fig:group-thresholds_celeba}
\end{figure*}

\pgfplotsset{compat=1.15}

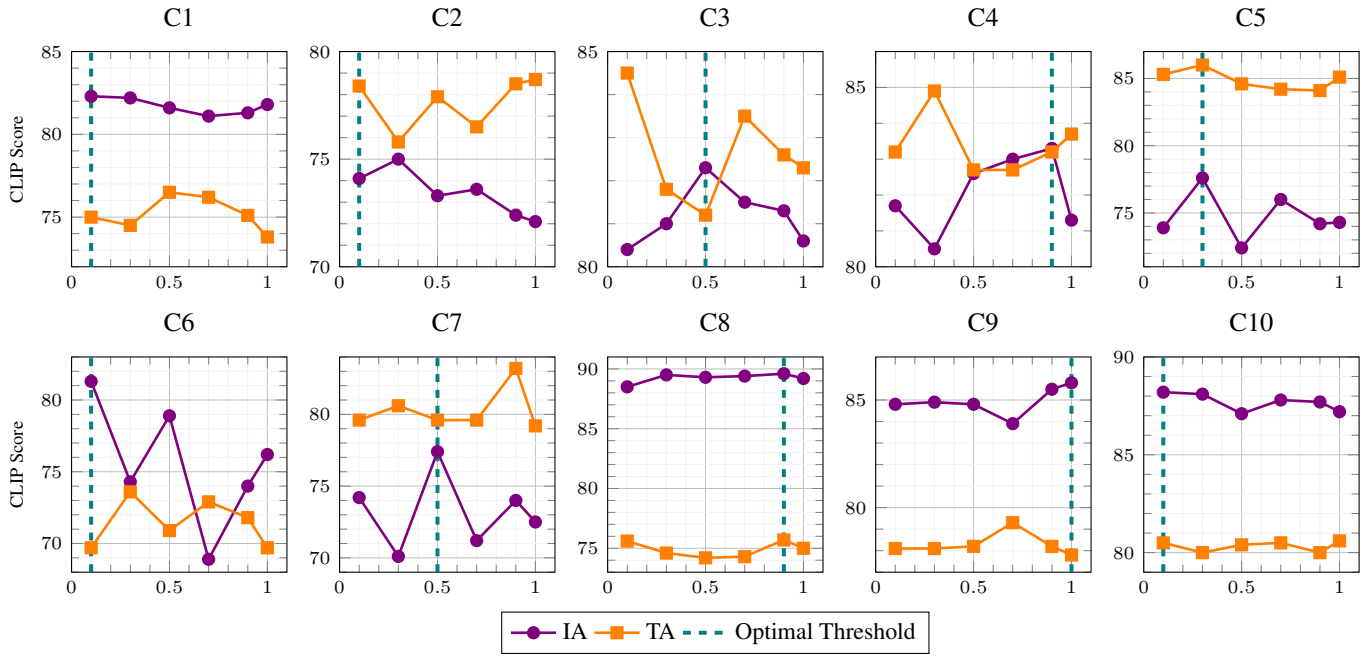
\begin{figure*}[!h]
\centering
\begin{tikzpicture}
\begin{groupplot}[
  group style={
    group size=5 by 2,
    horizontal sep=0.7cm,
    vertical sep=1.2cm,
  },
  width=0.25\textwidth,
  height=0.25\textwidth,
  xmin=0, xmax=1.1,
  ymin=55, ymax=90,
  ytick distance=5,
  label style={font=\scriptsize},
  tick label style={font=\scriptsize},
  grid=both,
  grid style={line width=.1pt, draw=gray!10},
  major grid style={line width=.2pt,draw=gray!50},
  minor tick num=4,
  legend style={font=\small},
]

\nextgroupplot[title=C1, ylabel={CLIP Score}, ymin=72, ymax=85, legend to name=zelda, legend columns =3]
\addplot[mark=*, violet, line width=1pt] table[x=Threshold, y=IA] {data_inet1.dat};
\addlegendentry{IA}
\addplot[mark=square*, orange, line width=1pt] table[x=Threshold, y=TA] {data_inet1.dat};
\addlegendentry{TA}
\addplot[
  dashed,
  thick,
  teal,
  line width = 1.5pt
] coordinates {(0.1, 72) (0.1, 85)};
\addlegendentry{Optimal Threshold}

\nextgroupplot[title=C2, ymin=70, ymax=80]
\addplot[mark=*, violet, line width=1pt] table[x=Threshold, y=IA] {data_inet2.dat};
\addplot[mark=square*, orange, line width=1pt] table[x=Threshold, y=TA] {data_inet2.dat};
\addplot[
  dashed,
  thick,
  teal,
  line width = 1.5pt
] coordinates {(0.1, 70) (0.1, 80)};

\nextgroupplot[title=C3, ymin=80, ymax=85]
\addplot[mark=*, violet, line width=1pt] table[x=Threshold, y=IA] {data_inet3.dat};
\addplot[mark=square*, orange, line width=1pt] table[x=Threshold, y=TA] {data_inet3.dat};
\addplot[
  dashed,
  thick,
  teal,
  line width = 1.5pt
] coordinates {(0.5, 80) (0.5, 85)};

\nextgroupplot[title=C4, ymin=80, ymax=86]
\addplot[mark=*, violet, line width=1pt] table[x=Threshold, y=IA] {data_inet4.dat};
\addplot[mark=square*, orange, line width=1pt] table[x=Threshold, y=TA] {data_inet4.dat};
\addplot[
  dashed,
  thick,
  teal,
  line width = 1.5pt
] coordinates {(0.9, 80) (0.9, 86)};

\nextgroupplot[title=C5, ymin=71, ymax=87]
\addplot[mark=*, violet, line width=1pt] table[x=Threshold, y=IA] {data_inet5.dat};
\addplot[mark=square*, orange, line width=1pt] table[x=Threshold, y=TA] {data_inet5.dat};
\addplot[
  dashed,
  thick,
  teal,
  line width = 1.5pt
] coordinates {(0.3, 71) (0.3, 87)};

\nextgroupplot[title=C6, ylabel={CLIP Score}, ymin=68, ymax=83]
\addplot[mark=*, violet, line width=1pt] table[x=Threshold, y=IA] {data_inet6.dat};
\addplot[mark=square*, orange, line width=1pt] table[x=Threshold, y=TA] {data_inet6.dat};
\addplot[
  dashed,
  thick,
  teal,
  line width = 1.5pt
] coordinates {(0.1, 68) (0.1, 83)};

\nextgroupplot[title=C7,ymin=69, ymax=84]
\addplot[mark=*, violet, line width=1pt] table[x=Threshold, y=IA] {data_inet7.dat};
\addplot[mark=square*, orange, line width=1pt] table[x=Threshold, y=TA] {data_inet7.dat};
\addplot[
  dashed,
  thick,
  teal,
  line width = 1.5pt
] coordinates {(0.5, 69) (0.5, 84)};

\nextgroupplot[title=C8, ymin=73, ymax=91]
\addplot[mark=*, violet, line width=1pt] table[x=Threshold, y=IA] {data_inet8.dat};
\addplot[mark=square*, orange, line width=1pt] table[x=Threshold, y=TA] {data_inet8.dat};
\addplot[
  dashed,
  thick,
  teal,
  line width = 1.5pt
] coordinates {(0.9, 73) (0.9, 91)};

\nextgroupplot[title=C9, ymin=77, ymax=87]
\addplot[mark=*, violet, line width=1pt] table[x=Threshold, y=IA] {data_inet9.dat};
\addplot[mark=square*, orange, line width=1pt] table[x=Threshold, y=TA] {data_inet9.dat};
\addplot[
  dashed,
  thick,
  teal,
  line width = 1.5pt
] coordinates {(1, 77) (1, 87)};

\nextgroupplot[title=C10, ymin=79, ymax=90]
\addplot[mark=*, violet, line width=1pt] table[x=Threshold, y=IA] {data_inet10.dat};
\addplot[mark=square*, orange, line width=1pt] table[x=Threshold, y=TA] {data_inet10.dat};
\addplot[
  dashed,
  thick,
  teal,
  line width = 1.5pt
] coordinates {(0.1, 79) (0.1, 90)};

\end{groupplot}
\path (group c1r2.south west) -- (group c5r2.south east) coordinate[midway] (centerbelow);
\node[anchor=north] at (centerbelow |- current bounding box.south) {\pgfplotslegendfromname{zelda}};
\end{tikzpicture}
\caption{\textbf{Effect of threshold variation on the ImageNet dataset.} In this plot, we analyse the variation of CLIP scores (\textcolor{violet}{\textbf{IA}} and \textcolor{orange}{\textbf{TA}}) for different threshold values. We observe that each concept has a specific threshold value that exhibits maximum gains in scores.}
\label{fig:group-thresholds_inet}
\end{figure*}

\subsection{Hyperbolic Constraint on Image Attention Maps vs. LoRA Weights}
\label{subsec:LoRA_v_img_attn}
In this section, we delve into the intuition behind our proposed approach and provide the reader with a compelling understanding of our results - quantitatively and qualitatively.

Most prior works on custom diffusion models in a continual learning setup propose solutions by limiting the inter-concept interference. Notably, our first experiment, ``CIDM with Cosine-guidance" on the CIFC dataset in Table \ref{tab:main_ablation} questions this strategy. We weight the Task Specific Loss ($\mathcal{L}_1$) based on the cosine-similarity scores between the token embeddings of the previous concepts and new concept, as shown in Eq. \ref{eq:ccl_weighted}. The comparable results of this experiment with ``CIDM" alone prompted us to explore the idea of leveraging inter-concept interactions rather than reducing them. Intuitively, we arrived at the goal of transferring learning from one concept (parent) to another (child). 

\begin{equation}
\label{eq:ccl_weighted}
\mathcal{L}_1 = \sum_{i=1}^{g-1} \sum_{l=1}^{L} \lambda_i \mathbf{A}_i^l (\mathbf{A}_g^l)^\top 
\end{equation}

Hyperbolic geometry allows us to exploit this potential relationship by naturally allowing embeddings to form tree-like hierarchies. This brings the question, why can't we directly optimize on the LoRA parameters in the hyperbolic space? Direct hyperbolic optimization on LoRA weights forces a simultaneous geometric transformation of both modalities. Although image-related weights adapt well to hyperbolic hierarchy preservation, text-related weights will suffer from distorted similarity metrics, breaking original semantic relationships \cite{modality_gap}. Thereby creating a zero-sum game where image gains come at text's expense, as shown in Table \ref{tab:LoRA_vs_imgAttn}.

To this end, we drive our focus on specifically only image attention maps of the incoming concept. Recall that our learning objective is designed to ensure that the attention maps of a child's concept lie in the entailment cone of the parent.

\begin{table*}[!ht]
    \centering
    \renewcommand{\arraystretch}{1.5}
    \resizebox{1\textwidth}{!}{
    \begin{tabular}{c|cc|cc|cc|cc|cc|cc|cc|cc|cc|cc|cccc}
        \hline
        \hline
        \multirow{2}{*}{\textbf{Methods}} & \multicolumn{2}{|c|}{\textbf{C1}} & \multicolumn{2}{|c|}{\textbf{C2}} & \multicolumn{2}{|c|}{\textbf{C3}} & \multicolumn{2}{|c|}{\textbf{C4}} & \multicolumn{2}{|c|}{\textbf{C5}} & \multicolumn{2}{|c|}{\textbf{C6}} & \multicolumn{2}{|c|}{\textbf{C7}} & \multicolumn{2}{|c|}{\textbf{C8}} & \multicolumn{2}{|c|}{\textbf{C9}} & \multicolumn{2}{|c|}{\textbf{C10}} & \multicolumn{2}{|c}{\textbf{Avg.}} \\
        \cline{2-23}  
         & \textbf{IA} &  \textbf{TA} & \textbf{IA} &  \textbf{TA} & 
         \textbf{IA} &  \textbf{TA} & 
         \textbf{IA} &  \textbf{TA} & 
         \textbf{IA} &  \textbf{TA} & 
         \textbf{IA} &  \textbf{TA} & 
         \textbf{IA} &  \textbf{TA} & 
         \textbf{IA} &  \textbf{TA} & 
         \textbf{IA} &  \textbf{TA} & 
         \textbf{IA} &  \textbf{TA} & 
         \textbf{IA} &  \textbf{TA} \\
         \hline

           \shortstack{\textbf{FLLP on} \\ \textbf{LoRA weights}} & 
           \textbf{84.3}&  73.3 & 
           \textbf{87.4}&  79.5 & 
           82.6&  \underline{73.1} & 
           83.2&  79.2 & 
           87.2&  75.7 & 
           69.1&  \underline{72.2} & 
           \textbf{84.5}&  71.8 & 
           \textbf{59.0}&  76.5 &
           \textbf{80.9}&  74.6 & 
           \textbf{77.3}&  69.5 & 
           79.6&  74.5\\

          \shortstack{\textbf{FLLP on image} \\ \textbf{attention maps}}& 
         83.9 &   \underline{75.7} & 
          87.0 &   \underline{82.3} & 
          \textbf{85.8} &   72.7 &  
           \textbf{83.5} &  \underline{82.7} & 
          \textbf{87.6} &   \underline{76.9} & 
          \textbf{72.2} &   72.1 & 
          84.2 &  \underline{72.5} & 
          58.8 &  \underline{77.5} & 
          80.2 &  \underline{77.3} & 
          76.4 &  \underline{71.6} & 
          \textbf{80.0} &  \underline{76.1} \\

          \rowcolor{lightgray}
         \textbf{$\Delta$} & 
          -0.4 & +2.4 & 
           -0.4&+2.8  & 
           +3.2 &-0.4  & 
            +0.3&+3.5  & 
           +0.4&+1.2  &
           +3.1&-0.1  & 
           -0.3&+0.7  & 
           -0.2&+1.0  & 
           -0.7&+2.7  & 
           -0.9&+2.1  & 
           +0.4&+1.6  \\
           \hline
           \hline
    \end{tabular}}
    \caption{\textbf{Hyperbolic constraint on LoRA weights instead of image attention maps.} We compare the effects of Hyperbolic Parent Entailment Loss directly on LoRA weights and image attention maps. Experimental results indicate that FLLP performs better when hyperbolic constraints are applied on the image attention maps. The higher IA score has been denoted in \textbf{bold}, and the higher TA value has been \underline{underlined}.}
    \label{tab:LoRA_vs_imgAttn}
\end{table*}

\begin{table*}[!ht]
    \centering
    \renewcommand{\arraystretch}{1.5}
    \resizebox{1\textwidth}{!}{
    \begin{tabular}{c|cc|cc|cc|cc|cc|cc|cc|cccc}
        \hline
        \hline
        \multirow{2}{*}{\textbf{Methods}} & \multicolumn{2}{|c|}{\textbf{C1 - C5}} & \multicolumn{2}{|c|}{\textbf{C6 - C10}} & \multicolumn{2}{|c|}{\textbf{C11 - C15}} & \multicolumn{2}{|c|}{\textbf{C16 - C20}} & \multicolumn{2}{|c|}{\textbf{C21 - C25}} & \multicolumn{2}{|c|}{\textbf{C26 - C30}} & \multicolumn{2}{|c|}{\textbf{C31 - C35}} & \multicolumn{2}{|c}{\textbf{Avg.}} \\
        \cline{2-17}  
         & \textbf{IA} &  \textbf{TA} & \textbf{IA} &  \textbf{TA} & 
         \textbf{IA} &  \textbf{TA} & 
         \textbf{IA} &  \textbf{TA} & 
         \textbf{IA} &  \textbf{TA} & 
         \textbf{IA} &  \textbf{TA} & 
         \textbf{IA} &  \textbf{TA} &  
         \textbf{IA} &  \textbf{TA} \\
         \hline

         \textbf{CIDM} \cite{cidm}  & 
          73.8 &   75.4& 
          70.2 &   75.9& 
          67.8 &   75.7 & 
          74.0 &  70.7 & 
          70.7 &   79.7 & 
          69.6 &   73.7& 
          70.8 &   78.2 &
          71.0 &   75.6 \\

          \textbf{FLLP}  & 
           \textbf{76.1}&  \underline{76.5} & 
           \textbf{71.7}&  \underline{77.2} & 
           \textbf{69.5}&  \underline{76.4} & 
           \textbf{75.9}&  \underline{71.4} & 
           \textbf{73.3}&  \underline{81.0} & 
           \textbf{71.9}&  \underline{74.7} & 
           \textbf{73.2}&  \underline{78.7} &  
           \textbf{73.1}&  \underline{76.6}\\

          \rowcolor{lightgray}
         \textbf{$\Delta$} & 
           +2.3& +1.1 & 
           +1.5& +1.3 & 
           +1.7& +0.7  & 
           +1.9&+0.7  & 
           +2.6&+1.3  &
           +2.3& +1.0 & 
           +2.4& +0.5 & 
           +2.1& +1.0 \\
           \hline
           \hline
    \end{tabular}}
    \caption{\textbf{Scalability Analysis.} To understand the robustness of FLLP on more than ten concepts, we consider 35 concepts (limited by CLIP tokenizer) from the CustomConcept101 dataset \cite{customconcept101}. We observe that FLLP consistently outperforms CIDM across both IA and TA scores (averaged over five consecutive concepts). The higher IA score has been denoted in \textbf{bold}, and the higher TA value has been \underline{underlined}.}
    \label{tab:scalability}
\end{table*}

\begin{table*}[!h]
    \centering
    \renewcommand{\arraystretch}{1.5}
    \resizebox{1\textwidth}{!}{
    \begin{tabular}{c|cc|cc|cc|cc|cc|cc|cc|cc|cc|cc|cccc}
        \hline
        \hline
        \multirow{2}{*}{\textbf{Methods}} & \multicolumn{2}{|c|}{\textbf{C1}} & \multicolumn{2}{|c|}{\textbf{C2}} & \multicolumn{2}{|c|}{\textbf{C3}} & \multicolumn{2}{|c|}{\textbf{C4}} & \multicolumn{2}{|c|}{\textbf{C5}} & \multicolumn{2}{|c|}{\textbf{C6}} & \multicolumn{2}{|c|}{\textbf{C7}} & \multicolumn{2}{|c|}{\textbf{C8}} & \multicolumn{2}{|c|}{\textbf{C9}} & \multicolumn{2}{|c|}{\textbf{C10}} & \multicolumn{2}{|c}{\textbf{Avg.}} \\
        \cline{2-23}  
         & \textbf{IA} &  \textbf{TA} & \textbf{IA} &  \textbf{TA} & 
         \textbf{IA} &  \textbf{TA} & 
         \textbf{IA} &  \textbf{TA} & 
         \textbf{IA} &  \textbf{TA} & 
         \textbf{IA} &  \textbf{TA} & 
         \textbf{IA} &  \textbf{TA} & 
         \textbf{IA} &  \textbf{TA} & 
         \textbf{IA} &  \textbf{TA} & 
         \textbf{IA} &  \textbf{TA} & 
         \textbf{IA} &  \textbf{TA} \\
         \hline
         
         \textbf{CIDM}\cite{cidm} & 
         \textbf{83.6} &  \underline{75.3} & 
         \textbf{86.4} &  78.1 & 
         \textbf{82.9} &  74.0 & 
         \textbf{80.8} &  81.1 & 
         \textbf{86.5} &  78.2 & 
         69.5 &  70.1 & 
         73.7 &  \underline{74.7} & 
         56.9 &  74.3 & 
         82.4 &  73.5 & 
         75.9 &  70.2 & 
         78.0 &  74.8 \\
         
         \shortstack{\textbf{CIDM} \cite{cidm} with \\ cosine-guidance}  & 
          82.0 &  74.4 & 
          85.0 &  \underline{80.1} & 
          81.1 &  \underline{75.5} & 
          78.3 &  \underline{81.6} & 
          85.4 &  \underline{79.3} & 
          \textbf{73.4} &  \underline{71.0} & 
          \textbf{81.9} &  72.1 & 
          \textbf{59.1} &  \underline{76.2} &
          \textbf{82.8} &  \underline{73.7} & 
          \textbf{76.9} &  \underline{71.8} & 
          \textbf{78.6} &  \underline{75.6}\\
          \rowcolor{lightgray}
         \textbf{$\Delta$} & 
           -1.6& -0.9 & 
           -1.4& +2.0  & 
           -1.8& +1.5 & 
           -2.5& +0.5 & 
           -1.1& +1.1 &
           +3.9& +0.9 & 
           +8.2& -2.6 & 
           +2.2& +1.9 & 
           +0.4& +0.2 & 
           +1.0& +1.6 & 
           +0.6& +0.8 \\
           \hline
           \hline
    \end{tabular}}
    \caption{\textbf{Intuition behind leveraging inter-concept interactions.} We compare the SOTA framework CIDM \cite{cidm} against CIDM with cosine-guidance that uses Eq. \ref{eq:ccl_weighted}. The weighting is derived based on cosine-similarity scores of the token embeddings. We observe an improvement in both Image Alignment (IA) and Text Alignment (TA), validating our hypothesis that inter-concept interactions can be utilized positively. The higher IA score has been denoted in \textbf{bold}, and the higher TA value has been \underline{underlined}.}
    \label{tab:main_ablation}
\end{table*}

\subsection{Computational Cost Analysis}
CIDM~\cite{cidm} and FLLP (Ours) exhibit comparable training times on the CIFC dataset - 4512s and 4910s, respectively. While hyperbolic operations in FLLP are inherently less efficient than their Euclidean counterparts, the overall training time is balanced since FLLP only considers a subset of prior concepts in its parent chain. Notably, inference time remains unchanged across both frameworks.

\subsection{Scalability Analysis}
To evaluate FLLP beyond ten concepts, we conduct experiments on the CustomConcept101 dataset~\cite{customconcept101}. FLLP builds upon CIDM~\cite{cidm}, which in turn uses the CLIP text encoder~\cite{clip}. Due to CLIP’s tokenizer constraint allowing only 77 additional tokens~\cite{clip}, we are limited to training on 35 concepts (considering 2–3 tokens per concept). Our results, as shown in Table \ref{tab:scalability}, indicate that FLLP scales well and consistently outperforms CIDM.

\subsection{Comparing Parameter Drifts}
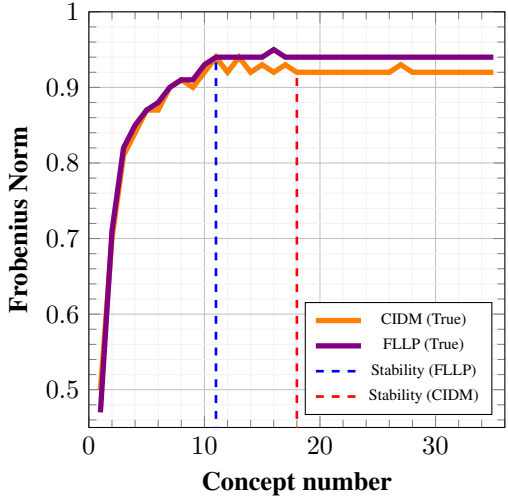
\begin{figure}[!h]
  \centering
  \begin{tikzpicture}
    \begin{axis}[
      width=0.4\textwidth,
      height=0.4\textwidth,
      xlabel={Concept number},
      ylabel={Frobenius Norm},
      font=\bfseries,
      legend pos=south east,
      legend style ={font=\small},
      xmin=0, xmax=36,
      ymin=0.45, ymax=1,
      grid=both,
      grid style={line width=.1pt, draw=gray!10},
      major grid style={line width=.2pt,draw=gray!50},
      minor tick num=4,
      legend style={font=\tiny}
    ]

    \addplot[orange, line width = 2pt] table[x=Num, y=Norm] {
      Num Norm
      1 0.5
      2 0.7
      3 0.81
      4 0.84
      5 0.87
      6 0.87
      7 0.9
      8 0.91
      9 0.9
      10 0.92
      11 0.94
      12 0.92
      13 0.94
      14 0.92
      15 0.93
      16 0.92
      17 0.93
      18 0.92
      19 0.92
      20 0.92
      21 0.92
      22 0.92
      23 0.92
      24 0.92
      25 0.92
      26 0.92
      27 0.93
      28 0.92
      29 0.92
      30 0.92
      31 0.92
      32 0.92
      33 0.92
      34 0.92
      35 0.92
    };
    \addlegendentry{CIDM (True)}

    \addplot[violet, line width = 2pt] table[x=Num, y=Norm] {
      Num Norm
      1 0.47
      2 0.71
      3 0.82
      4 0.85
      5 0.87
      6 0.88
      7 0.9
      8 0.91
      9 0.91
      10 0.93
      11 0.94
      12 0.94
      13 0.94
      14 0.94
      15 0.94
      16 0.95
      17 0.94
      18 0.94
      19 0.94
      20 0.94
      21 0.94
      22 0.94
      23 0.94
      24 0.94
      25 0.94
      26 0.94
      27 0.94
      28 0.94
      29 0.94
      30 0.94
      31 0.94
      32 0.94
      33 0.94
      34 0.94
      35 0.94
    };
    \addlegendentry{FLLP (True)}

    \addplot[dashed, blue, line width = 1pt] table[x=Num, y=Norm] {
      Num Norm
      11 0
      11 0.47
      11 0.5
      11 0.6
      11 0.8
      11 0.94
    };
    \addlegendentry{Stability (FLLP)}

    \addplot[dashed, red, line width = 1pt] table[x=Num, y=Norm] {
      Num Norm
      18 0
      18 0.47
      18 0.5
      18 0.6
      18 0.8
      18 0.92
    };
    \addlegendentry{Stability (CIDM)}

    \end{axis}
  \end{tikzpicture}
  \caption{\textbf{Comparing True LoRA Frobenius Norms.} Both frameworks exhibit an upward trajectory in their norm values, suggesting significant updates to the parameters with incoming concepts. FLLP is able to incorporate concepts without parameter updates earlier than CIDM, as evidenced by their stability points indicated by \textcolor{blue}{blue} and \textcolor{red}{red} lines, respectively. }
  \label{fig:param_drift1}
\end{figure}

To study the evolution of LoRA weights in FLLP (Ours) and CIDM \cite{cidm}, we plot the Frobenius norm of the LoRA weights averaged across all layers against 35 concepts from the CustomConcept101 \cite{customconcept101} dataset as shown in Fig. \ref{fig:param_drift1}. Both CIDM and FLLP show a clear trend in the Frobenius norm values of their LoRA parameters over the 35 concepts. Initially, both frameworks exhibit a rapid increase in their norm values, suggesting significant updates or adjustments to the parameters as learning progresses. For CIDM, the values start at 0.5 and climb to around 0.92-0.94 by step 19, where they stabilize with minor fluctuations. Similarly, FLLP begins at 0.47 and rises to 0.94-0.95 by step 11, stabilizing thereafter. This indicates that FLLP is able to incorporate concepts without significant parameter updates earlier than CIDM. Using the relative Frobenius norm as a measure, we calculated the average absolute difference between consecutive steps to quantify the parameter drift for each framework in Fig. \ref{fig:param_drift2}. \textbf{For CIDM, the average drift is 0.018}. \textbf{FLLP exhibits a 22\% lower average drift at 0.014}, suggesting even less variation per concept on average compared to CIDM. Thereby exhibiting superior performance in addressing catastrophic forgetting.

\begin{figure}[!h]
  \centering
  \begin{tikzpicture}
    \begin{axis}[
      width=0.4\textwidth,
      height=0.4\textwidth,
      xlabel={Concept number},
      ylabel={Relative Frobenius Norm},
      font=\bfseries,
      legend pos=north east,
      legend style ={font=\small},
      xmin=0, xmax=36,
      ymin=-0.01, ymax=0.25,
      grid=both,
      grid style={line width=.1pt, draw=gray!10},
      major grid style={line width=.2pt,draw=gray!50},
      minor tick num=4,
      legend style={font=\tiny}
    ]

    \addplot[orange, line width = 2pt] table[x=Num, y=Norm] {
      Num Norm
      2 0.2
      3 0.11
      4 0.03
      5 0.03
      6 0
      7 0.03
      8 0.01
      9 0.01
      10 0.02
      11 0.02
      12 0.02
      13 0.02
      14 0.02
      15 0.01
      16 0.01
      17 0.01
      18 0.01
      19 0
      20 0
      21 0
      22 0
      23 0
      24 0
      25 0
      26 0
      27 0.01
      28 0.01
      29 0.01
      30 0.01
      31 0
      32 0
      33 0
      34 0
      35 0
    };
    \addlegendentry{CIDM (Relative)}

    \addplot[violet, line width = 2pt] table[x=Num, y=Norm] {
      Num Norm
      2 0.24
      3 0.11
      4 0.03
      5 0.02
      6 0.01
      7 0.02
      8 0.01
      9 0
      10 0.02
      11 0.01
      12 0
      13 0
      14 0
      15 0.01
      16 0
      17 0.01
      18 0
      19 0
      20 0
      21 0
      22 0
      23 0
      24 0
      25 0
      26 0
      27 0
      28 0
      29 0
      30 0
      31 0
      32 0
      33 0
      34 0
      35 0
    };
    \addlegendentry{FLLP (Relative)}
    \end{axis}
  \end{tikzpicture}
  \caption{\textbf{Comparing Relative LoRA Frobenius Norms.}} We compute the difference between LoRA Frobenius Norms as a metric to measure parameter drift. Based on this experiment, we observe that FLLP exhibits a 22\% lower average parameter drift compared to CIDM, thereby showcasing its advantages in alleviating catastrophic forgetting. 
  \label{fig:param_drift2}
\end{figure}

\subsection{Understanding Catastrophic Forgetting in a 1D Synthetic Dataset}

\paragraph{Experimental Setup.} The UNet is first trained on $1000$ samples for $\mu = 1$ and $\sigma = 1$. Subsequently, the UNet parameters are frozen, and only task-specific token embeddings are learned. During training, the model sequentially encounters 1000 samples for each task \(\mu_i = 3i, \text{where } i \in [1, 5]\), mimicking a continual learning setting. After training on all tasks, we evaluate the model’s retention of past concepts by comparing the generated mean of the reconstructed distribution to its true mean. The sum of the absolute differences between the generated and true means acts as a quantitative indicator of forgetting, termed as \textit{forgetting rate}: smaller differences suggest better retention and lower forgetting. 

We measure \textit{forgetting rate} across three models using this setup: baseline continual learning, CIDM \cite{cidm}, and FLLP (Ours). The baseline model is optimized only on Mean Squared Error (MSE), whereas CIDM uses a combination of MSE and Concept Consolidation Loss (CCL) \cite{cidm}. We do not consider other loss functions since CCL is the point of difference in the learning objective between CIDM and FLLP (Ours).

\subsection{Hyperbolic Parent Entailment Loss}

\begin{algorithm}[!h]
\caption{\textsc{Hierarchical Parent Entailment Loss}}
\label{alg:entailment_learning}
\begin{algorithmic}
\Require{ $X_{\text{new}}$ (image attention maps of new concept); $X_{\text{old}}$(image attention maps of previous concepts); $F_{\text{img}}$ (CLIP image features of new concept); $F_{\text{text}}$ (previous concepts' token embeddings);
$t$ (aperture threshold)
}
\\

\State $Z_{\text{new}}, Z_{\text{old}} \gets \texttt{expm}(X_\text{new}), \texttt{expm}(X_\text{old})$
\State $Z_{\text{img}}, Z_{\text{text}} \gets \texttt{expm}(F_\text{img}), \texttt{expm}(F_\text{text})$
\State $P \gets \texttt{ParentSearch}(Z_{\text{img}}, Z_{\text{text}})$
\Comment{Selects parents by iteratively moving up the tree}
\State $Loss \gets [\ ]$
\For{each index $i$ in $P$}
    \State $Z_p \gets Z_{\text{old}}[i]$
    \State $\theta \gets \texttt{Ext}(Z_p, Z_{\text{new}})$
    \Comment{Eq. \ref{eq:ext}}
    \State $\alpha \gets \texttt{Aper}(Z_p)$
    \Comment{Eq. \ref{eq:aperture}}
    \State $Loss.\texttt{append}(\max(0, \theta - t\alpha))$
\EndFor
\State \Return $\texttt{mean}(Loss)$
\end{algorithmic}
\end{algorithm}

To ensure hierarchical consistency in concept learning, we propose the Hyperbolic Parent Entailment Loss (Algorithm~\ref{alg:entailment_learning}), which encourages new image attention maps to respect parent-child entailment structure. The algorithm operates by first exponentiating both the incoming concept’s image attention maps and the prior concepts’ maps to obtain their hyperbolic embeddings. These embeddings are then matched to their respective parent candidates using a tree traversal procedure (\texttt{ParentSearch}) guided by the CLIP-derived image and text feature spaces. For each identified parent, we extract the entailment angle $\theta$ between the parent’s and the new concept’s embeddings, and weigh it against the parent’s aperture $\alpha$, its effective generalization radius. The loss penalizes violations of the entailment condition, i.e., when $\theta$ exceeds the scaled aperture $t\alpha$, ensuring that children remain geometrically constrained within the conceptual cone of their parents. This loss formulation allows for selective, hierarchy-preserving updates to the embedding space without regressing previously learned semantic structure.

\subsection{ParentSearch Algorithm}
\begin{algorithm}[h]
\caption{\textsc{ParentSearch}: Parent Chain Construction in Hyperbolic Space through Union-Find Algorithm}
\label{alg:parent-search}
\begin{algorithmic}[1]
\Require{$Z_{\text{img}}$ (new concept embedding); $Z_{\text{old}}$ (old concepts' embeddings)}
\State $\text{current\_emb} \gets \text{new\_emb}$
\State $\text{parent\_chain}, \text{visited} \gets \emptyset, \emptyset$
\State $\text{visited} \gets \emptyset$
\State $\text{search\_pool} \gets \text{old\_embs} \cup \text{new\_emb}$
\State $\text{index\_list} \gets [0, 1, \dots, n-1] \cup [-1]$

\While{True}
    \State $D \gets -d_{\mathcal{L}}(\text{current\_emb}, \text{search\_pool})$
    \Comment{Eq. \ref{eq:geodesic_dist}}
    
    \State $ \text{top\_indices} \gets \text{TopK}(D, \text{k=2}, \texttt{min=True})$
    \Comment{Pick the smallest 2}
    \State $\text{parent} \gets \text{top\_indices}[1]$
    \Comment{Since the smallest will be the embedding itself}
    
    \State $\text{parent\_idx} \gets \text{index\_list}[\text{parent}]$
    
    \If{$(\text{parent\_idx} \in \text{visited}) \lor (\text{parent\_idx} = -1)$}
    \Comment{A loop is detected}
        \State \textbf{break}
    \EndIf
    
    \State $\text{visited} \gets \text{visited} \cup \{\text{parent\_idx}\}$
    \State $\text{parent\_chain} \gets \text{parent\_chain} \cup \{\text{parent\_idx}\}$
    \State $\text{current\_emb} \gets \text{search\_pool}[\text{parent\_idx}]$
\EndWhile

\State \Return $\text{parent\_chain}$
\end{algorithmic}
\end{algorithm}

The ParentSearch algorithm (Algorithm~\ref{alg:parent-search}) constructs a hierarchical parent chain for a given new concept by performing iterative nearest-neighbor traversal in hyperbolic space. Leveraging the geodesic distance metric $d_{\mathcal{L}}$ in the Lorentz model, it identifies the closest ancestor candidates among all previously embedded concepts. To prevent cyclic dependencies and redundant loops, a union-find–style visitation set is maintained, ensuring that traversal proceeds only through unvisited nodes. At each step, the nearest neighbor (excluding the current embedding itself) is selected as the next parent, and its index is appended to the parent chain. This walk continues until a stopping criterion is met, either a previously visited node is encountered, or the traversal reaches the new embedding itself. The resulting chain serves as the foundation for enforcing entailment constraints, allowing each child to align itself within the conceptual envelope of its semantic predecessors. This mechanism elegantly encodes implicit tree structures without the need for explicit supervision or symbolic hierarchies. The algorithm has a worst-case time complexity of $O(c-1)$ when training for the $c$-th concept.

\subsection{Experiments, Implementation Details and Evaluation Metrics}
\label{subsec:appendix_evalDetails}
To showcase the robustness and generalizability of our model, we have used three public datasets: CIFC \cite{cidm}, CelebA \cite{celeba}, and ImageNet \cite{imagenet}. Each dataset consists of 10 different concepts with 3-5  reference images for each task. We use Stable Diffusion (SD-1.5) \cite{ldm} as the pretrained model and the CIDM \cite{cidm} as the foundational framework for all experiments. The training is conducted on two Nvidia H100 GPUs, with a fixed initial learning rate of \(1e^{-3}\) for updating textual embeddings and \( 1e^{-4}\) for optimizing the U-Net. The curvature of the hyperboloid is a learnable parameter initialized to $1.0$, and optimized with a learning rate of $1e-4$. Empirically, we set \(\gamma_1 = \gamma_2 = 0.1\) in Eq. \ref{eq:total_loss}.

Following our experiments under our problem setting as in Sec \ref{sec:problem}, we evaluate our generated images across 2 metrics - Image Alignment (IA) and Text Alignment (TA). Image Alignment (IA) scores are computed using the CLIP \cite{clip} image encoder, comparing the similarity of features between generated images and reference images. Similarly, we utilize the text encoder of CLIP \cite{clip} to evaluate the text-image similarity between the input prompt and synthesized image for the Text Alignment (TA) scores. To analyze the comparisons between our model and SOTA frameworks, we follow the evaluation strategy defined in \cite{cidm}. Specifically, 20 evaluation prompts are introduced for each of the 10 concepts, and 50 images are generated per prompt.

\subsection{Background into Diffusion Models}
\label{subsec:prelim}
Latent diffusion models (LDMs) \cite{ldm} rely on conditional inputs, such as text prompts \cite{ldm_text1} or images \cite{ldm_img2}, to guide the generation of images, utilizing an encoder $\Phi(\cdot)$ and a decoder $\Psi(\cdot)$. Custom diffusion models (CDMs) \cite{cdm1} extend LDMs by incorporating low-rank adaptation (LoRA) \cite{lora2} to fine-tune pretrained diffusion models \cite{ldm} for personalized concept learning.

Given a personalized image-text pair $(x, p)$, the encoder $\Phi(\cdot)$ maps $x$ to a latent representation $z$, with $z_t$ denoting the noisy latent feature at timestep $t$ ($t = 1, \dots, T$). The text encoder $\Gamma(\cdot)$ maps the text prompt $p$ to a textual embedding $c = \Gamma(p)$. The objective for learning a personalized concept ${(x, p)}$ at timestep $t$ is defined as:

\begin{equation}\label{eq:1}
    \mathcal{L}_{CDM} = \mathbb{E}_{z \sim \Phi(x), c, \varepsilon \sim \mathcal{N}(0,I), t} \left[ \| \varepsilon - \varepsilon_{\theta'}(z_t | c, t) \|_2^2 \right]
\end{equation}

where $\varepsilon_{\theta'}(\cdot)$ represents the denoising UNet \cite{ldm, ldm_text1} that gradually denoises $z_t$ by estimating the Gaussian noise $\varepsilon \sim \mathcal{N}(0, I)$. The parameter set $\theta'$ corresponds to  $\theta' = \theta_0 + \Delta\theta$, where $\theta_0 = \{W_l^0\}_{l=1}^{L}$ denotes the pretrained weights in LDMs, and $\Delta\theta = \{\Delta W_l\}_{l=1}^{L}$ corresponds to the LoRA-updated parameters \cite{mix_of_show, customconcept101}. Here, $W_l^0, \Delta W_l \in \mathbb{R}^{a \times b}$ are the pretrained and low-rank weight matrices in the $l$-th transformer layer of $\theta'$, respectively, where $a$ and $b$ are matrix dimensions. Following \cite{dreambooth}, the low-rank update $\Delta W_l$ can be factorized as $\Delta W_l = A_l B_l$,
where $A_l \in \mathbb{R}^{a \times r}$ and $B_l \in \mathbb{R}^{r \times b}$ with rank $r \ll \min(a, b)$. Note that the term “concept” refers to introducing the model to a new visual category, such as teaching it what a “zebra” looks like, as illustrated in Fig. \ref{fig:intro_hb_space}.

\subsection{Further Background into Hyperbolic Geometry}
\label{subsec:hyperbolic_bg}

\paragraph{Manifolds and Smooth Structures.} A \emph{manifold} is a topological space that is locally homeomorphic to Euclidean space. More precisely, an \( n \)-dimensional manifold \( M \) is a space where each point \( x \in M \) has a neighborhood that can be mapped bijectively and continuously (with a continuous inverse) to an open subset of \( \mathbb{R}^n \). This structure allows for the analysis of complex, curved spaces using local coordinates. When the transition maps between overlapping coordinate charts are differentiable, the manifold is called a \emph{smooth manifold}. Smooth manifolds provide the foundational framework for modern differential geometry, enabling calculus to be performed on non-Euclidean spaces.

\paragraph{Riemannian Manifolds.} A \emph{Riemannian manifold} \( (M, g) \) is a smooth manifold \( M \) equipped with a smoothly varying inner product \( g_x \) on the tangent space \( T_xM \) at each point \( x \in M \). The Riemannian metric \( g \) enables the measurement of geometric quantities such as the lengths of curves, angles between vectors, and notions of local curvature. The metric induces a distance function on the manifold, turning it into a metric space. Euclidean space \( \mathbb{R}^n \) is a simple example where the metric is constant and flat. In contrast, on general Riemannian manifolds, the curvature may vary spatially and fundamentally influences the geometry and topology of the space.

\paragraph{Hyperbolic Geometry.}Riemannian manifolds with constant negative curvature are called \emph{hyperbolic manifolds}. These spaces exhibit properties that sharply contrast with those of flat (Euclidean) or positively curved (spherical) manifolds, such as exponential volume growth and distinct geodesic behavior. The canonical example is hyperbolic space \( \mathbb{H}^n \), which can be realized through various models. The \emph{Poincar\'e ball model} maps \( \mathbb{H}^n \) to the open unit ball in \( \mathbb{R}^n \), preserving angles while distorting distances. Alternatively, the \emph{Lorentz model} (also called the hyperboloid model) embeds \( \mathbb{H}^n \) as a submanifold of \( \mathbb{R}^{n+1} \) equipped with a Lorentzian inner product. This model is particularly useful in machine learning due to its closed-form geodesic computations and efficient optimization properties.

\paragraph{Lorentzian inner product: } Let \( \langle \cdot, \cdot \rangle \) denote the standard Euclidean inner product in \( \mathbb{R}^n \). The \textit{Lorentzian inner product} for vectors \( x, y \in \mathbb{R}^{n+1} \), is defined as: \(\langle x, y \rangle_{\mathcal{L}} = \langle \mathbf{x}_{\text{space}}, \mathbf{y}_{\text{space}} \rangle - x_{\text{time}} y_{\text{time}}\).
This inner product induces the Lorentzian norm \( \| x \|_{\mathcal{L}} = \sqrt{|\langle x, x \rangle_{\mathcal{L}}|} \). The Lorentz model of hyperbolic space with constant negative curvature \( k \) is then given as : $\mathcal{L}^n_k = \left\{ x \in \mathbb{R}^{n+1} : \langle x, x \rangle_{\mathcal{L}} = -\frac{1}{k},\ x_{\text{time}} > 0 \right\})$. Each point \( x \in \mathcal{L}^n_k \) satisfies the constraint, $ x_{\text{time}} = \sqrt{\frac{1}{k} + \| \mathbf{x}_{\text{space}} \|^2}$.

\paragraph{Geodesics.} The shortest path between two points on a Riemannian manifold is defined as a \textit{geodesic}. In the Lorentz model of hyperbolic space, geodesics are defined as the intersections between the hyperboloid and hyperplanes that pass through the origin of the ambient space $\mathbb{R}^{n+1}$. The Lorentzian distance between two points $x, y \in \mathcal{L}^n$ is given by, 
\begin{equation}
\label{eq:geodesic_dist}
d_{\mathcal{L}}(x, y) = \frac{1}{\sqrt{k}} \cosh^{-1}(-k \langle x, y \rangle_{\mathcal{L}})
\end{equation}

\paragraph{Exponential and Logarithmic Maps.} The \textit{exponential map} at point $z$ maps tangent vectors $v \in T_z \mathcal{L}^n$ onto the manifold $\mathcal{L}^n$, describing how the tangent space folds back onto the manifold. Its inverse, the \textit{logarithmic map}, projects a point $x \in \mathcal{L}^n$ back to the tangent space.
\begin{equation}
\texttt{expm}_z(v) = \cosh(\sqrt{k} \|v\|_{\mathcal{L}}) z + \frac{\sinh(\sqrt{k} \|v\|_{\mathcal{L}})}{\sqrt{k} \|v\|_{\mathcal{L}}} v
\end{equation}
\begin{equation}
\texttt{logm}_z(x) = \frac{\cosh^{-1}(-k\langle z, x \rangle_{\mathcal{L}})}{\sqrt{(k \langle z, x \rangle_{\mathcal{L}})^2 - 1}} \text{proj}_z(x)
\end{equation}

In this paper, we use the origin $O = [\mathbf{0}, \sqrt{1/k}]$ of the hyperboloid as the reference point for both maps.

\paragraph{Tangent Space.} At any point $z \in \mathcal{L}^n$, the \textit{tangent space} $T_z \mathcal{L}^n$ forms a Euclidean subspace of $\mathbb{R}^{n+1}$ and contains all vectors orthogonal to $z$ under the Lorentzian inner product:
\begin{equation}
T_z \mathcal{L}^n = \{ v \in \mathbb{R}^{n+1} \mid \langle z, v \rangle_{\mathcal{L}} = 0 \}
\end{equation}

\subsection{Datasets}

We evaluate our approach on three distinct datasets, each curated to ensure high semantic clarity and alignment with our objectives. For the CIFC dataset, we directly adopt the benchmark provided in the reference paper, which includes conceptually rich visual cards that serve as a challenging ground for hallucination detection. For the ImageNet subset, we manually select 3–5 images per class across 20 classes, ensuring that the primary object of interest is prominently visible, well-centered, and unobstructed in each image. This careful selection mitigates background noise and ambiguous visual cues that could mislead the model during training or evaluation. Finally, for the CelebA dataset, we extract a subset comprising approximately 10 distinct identities, choosing 3–5 representative images per identity. Selection criteria included clear frontal facial orientation, consistent lighting, and minimal occlusion to maintain identity coherence across samples. This hand-picked curation across datasets ensures high-quality supervision signals, especially important in training models to reduce semantic hallucinations and enforce visual-textual alignment.

\begin{figure*}[!h]
    \centering
    \includegraphics[width=1\linewidth]{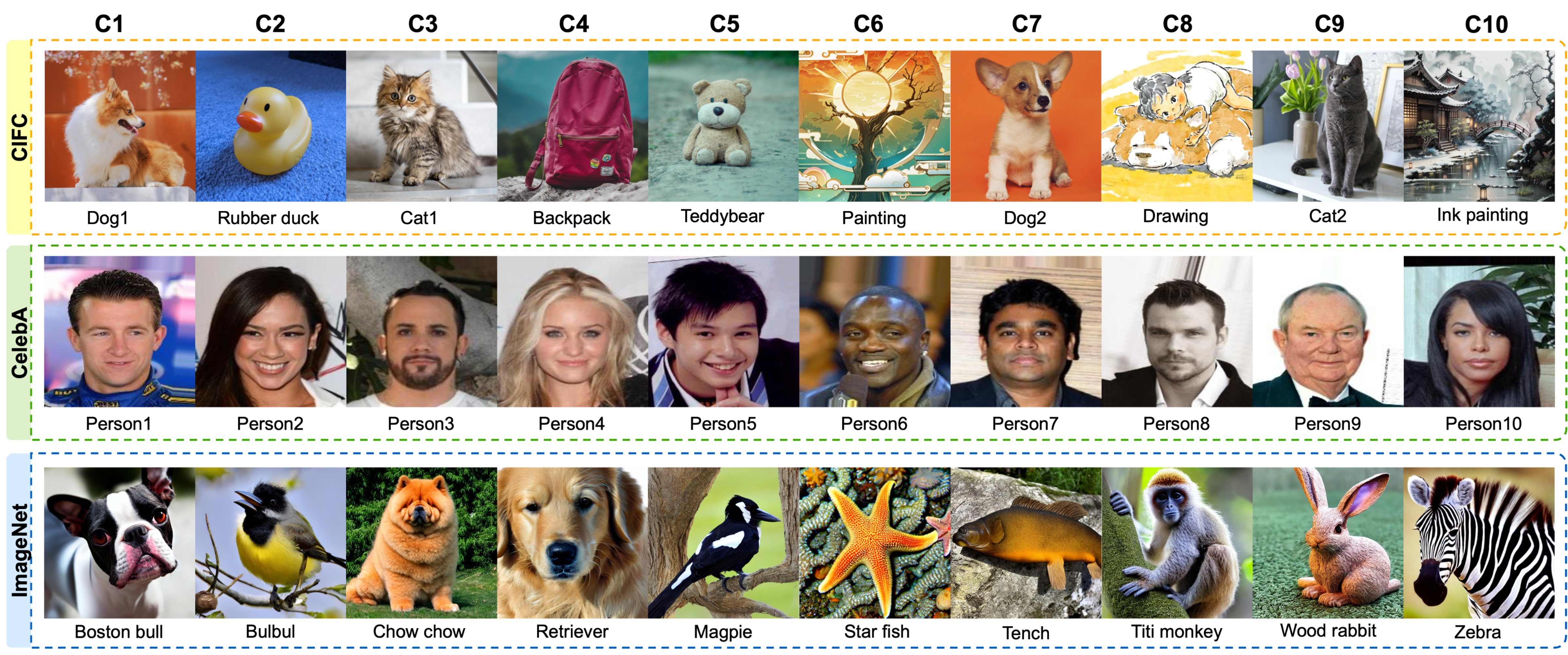}
    \caption{\textbf{Datasets.} Class-wise representative images from the datasets used in our experiments.}
    \label{fig:datasets}
\end{figure*}

\subsection{More Qualitative Results}

\begin{figure*}[!h]
    \centering
    \includegraphics[width=0.95\linewidth]{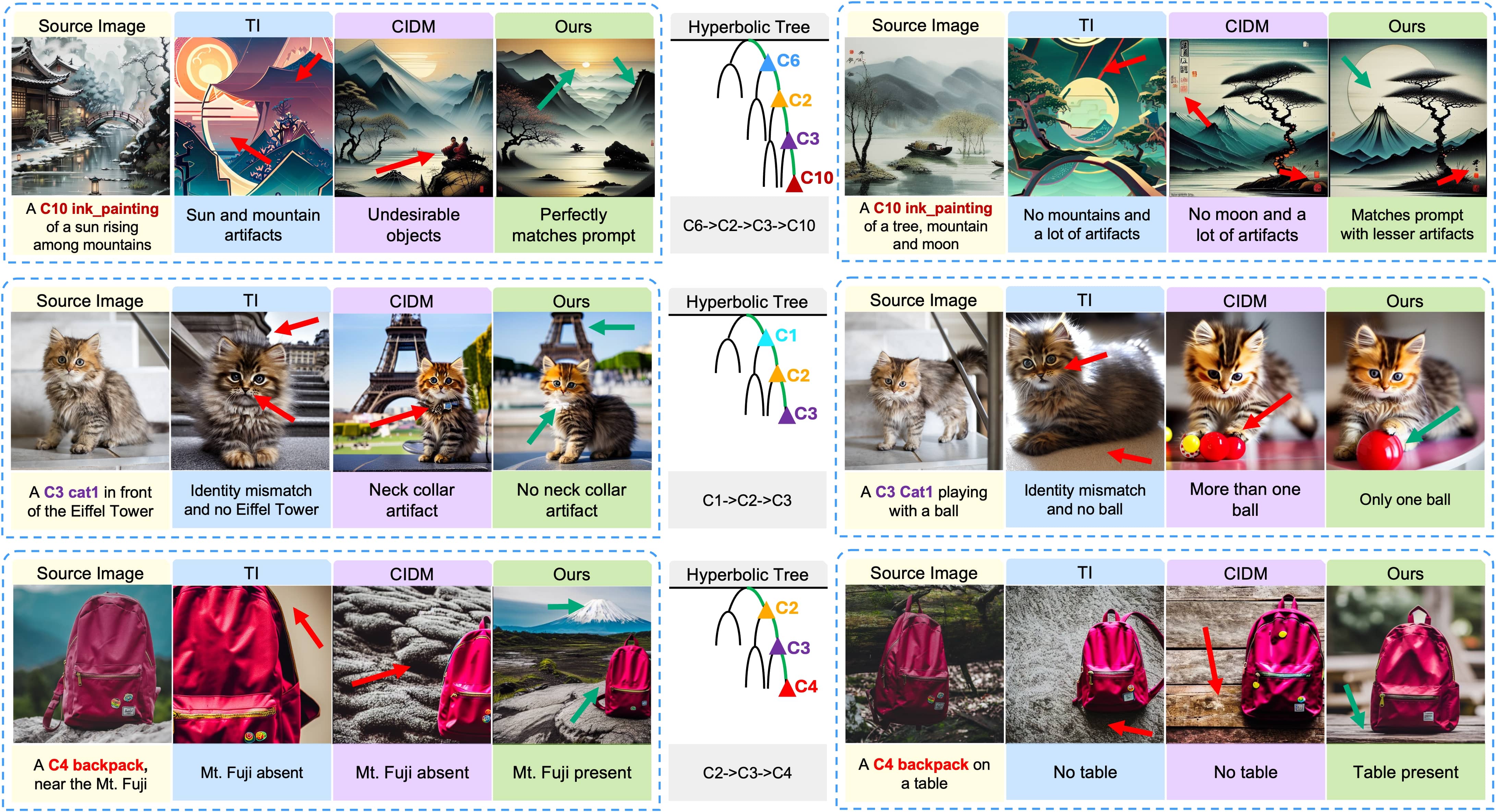}
    \caption{\textbf{Qualitative results on the CIFC dataset.} We compare the generated images in TI \cite{dosovitskiy2021imageworth16x16words}, CIDM \cite{cidm} and FLLP (Ours). The red and green arrows indicate regions of undesirable and desirable qualities, and their reasons are stated below each image. The hyperbolic tree indicates the parent chain of concepts that the model traversed in learning the incoming concept.}
    \label{fig:suppl_qual_cifc}
\end{figure*}

\begin{figure*}[!h]
    \centering
    \includegraphics[width=0.95\linewidth]{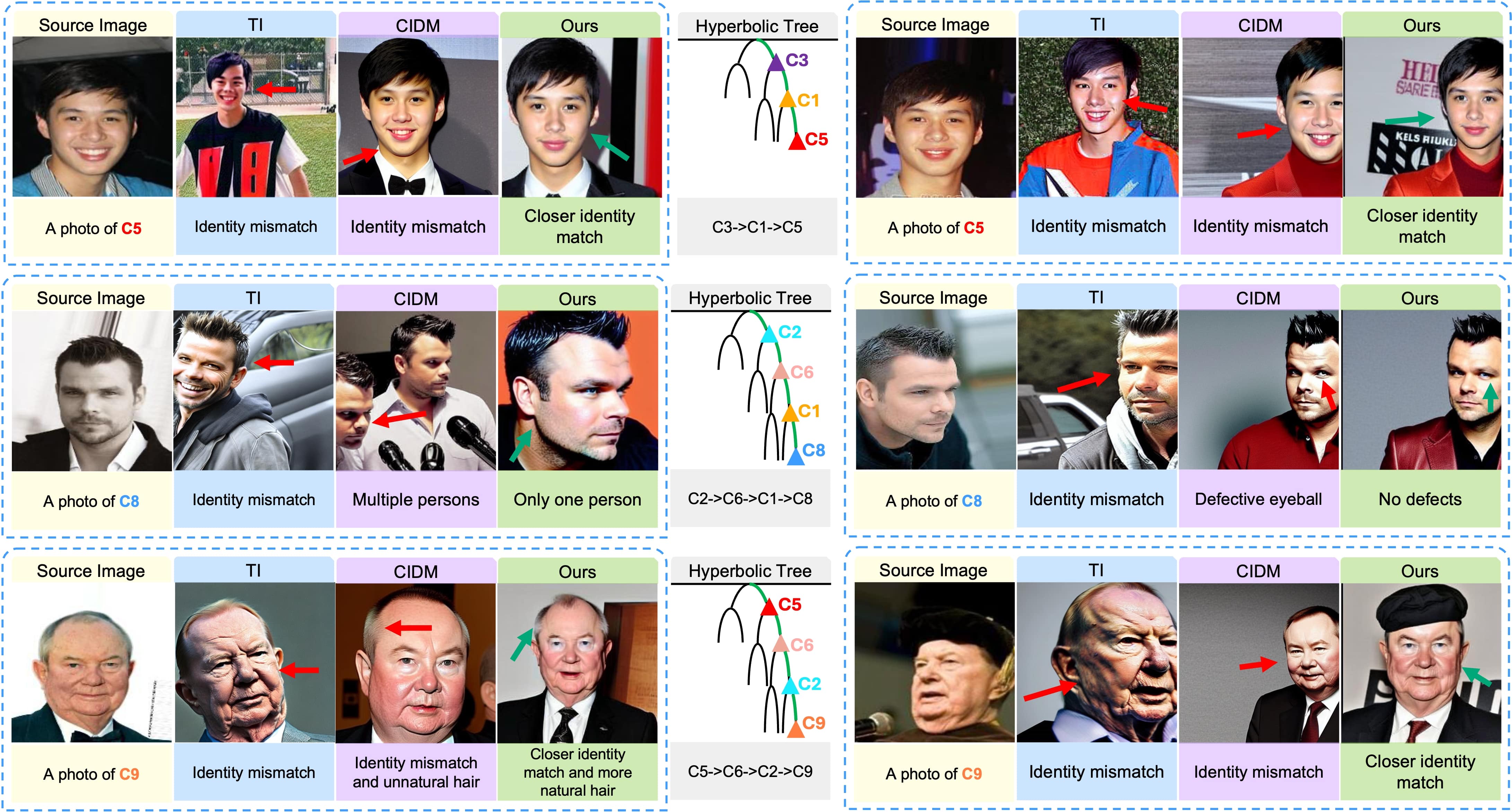}
    \caption{\textbf{Qualitative results on the CelebA dataset.} We compare the generated images in TI \cite{dosovitskiy2021imageworth16x16words}, CIDM \cite{cidm} and FLLP (Ours). The red and green arrows indicate regions of undesirable and desirable qualities, and their reasons are stated below each image. The hyperbolic tree indicates the parent chain of concepts that the model traversed in learning the incoming concept.}
    \label{fig:suppl_qual_celeba}
\end{figure*}

\begin{figure*}[!h]
    \centering
    \includegraphics[width=0.95\linewidth]{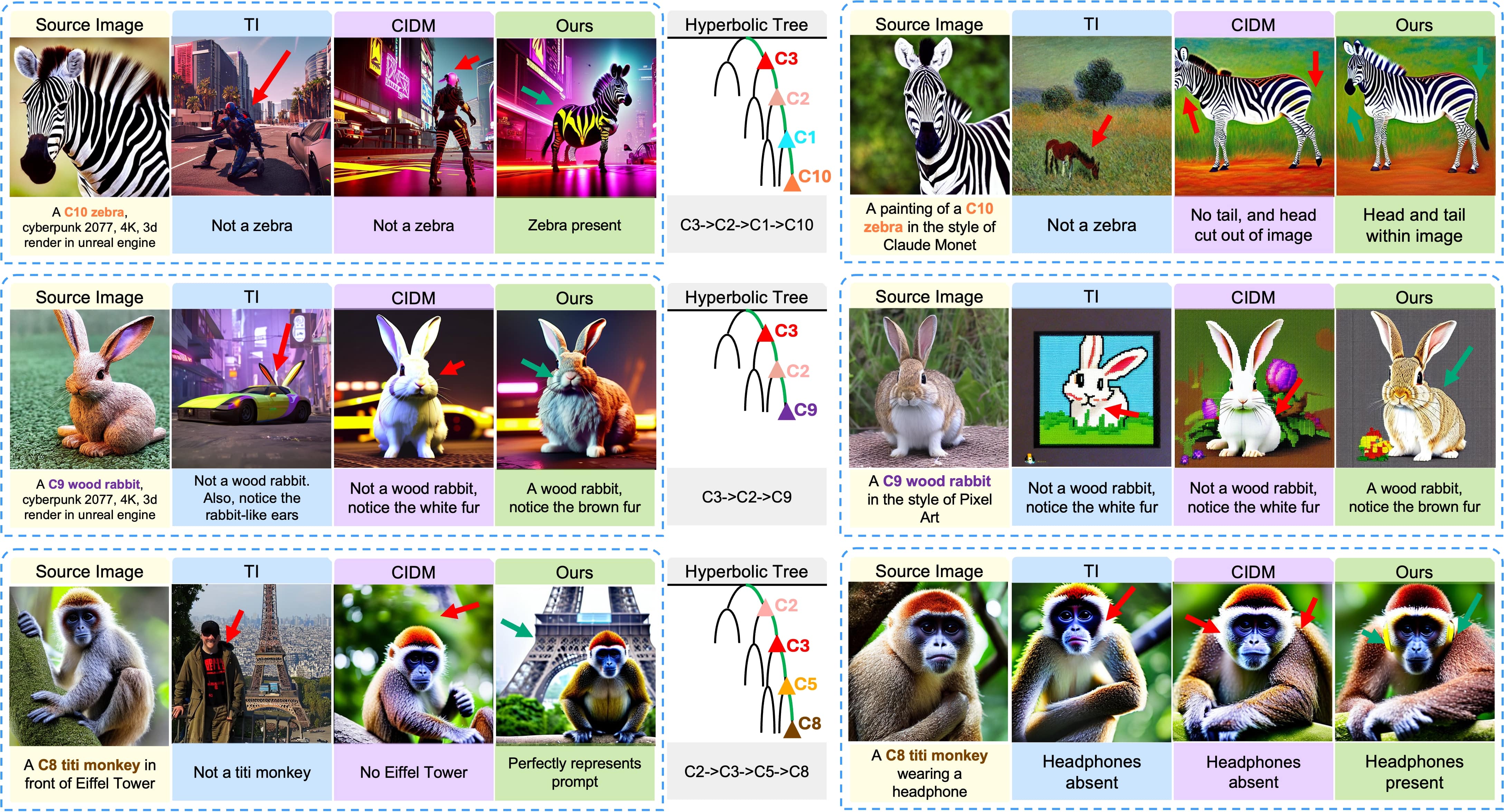}
    \caption{\textbf{Qualitative results on the ImageNet dataset.} We compare the generated images in TI \cite{dosovitskiy2021imageworth16x16words}, CIDM \cite{cidm} and FLLP (Ours). The red and green arrows indicate regions of undesirable and desirable qualities, and their reasons are stated below each image. The hyperbolic tree indicates the parent chain of concepts that the model traversed in learning the incoming concept.}
    \label{fig:suppl_qual_inet}
\end{figure*}

Figures~\ref{fig:suppl_qual_cifc}, \ref{fig:suppl_qual_celeba}, and \ref{fig:suppl_qual_inet} present additional qualitative comparisons of generations from FLLP (Ours) against prior methods on CIFC~\cite{cidm}, CelebA~\cite{celeba}, and ImageNet~\cite{imagenet}, respectively. Across all datasets, we observe consistent failure modes in TI~\cite{dosovitskiy2021imageworth16x16words} and CIDM~\cite{cidm}, notably the loss of subject identity and the emergence of extraneous or semantically irrelevant artifacts—often appearing in multiplicity. In contrast, FLLP (Ours) demonstrates stronger identity preservation and significantly reduced artifact generation, yielding image outputs with more desirable traits.

\subsection{Limitations}
Although CLIP \cite{clip} is a relatively older evaluation metric, it remains prevalent due to its computational efficiency, annotation-free nature, and scalability to large datasets. We recognize its limitations—particularly its reliance on global image-text similarity, which may fail to capture fine-grained or personalized attributes (e.g., ``a \textit{dirty} car", ``a \textit{yellow} tree"). As shown in Figures~\ref{fig:suppl_qual_cifc},~\ref{fig:suppl_qual_celeba}, and~\ref{fig:suppl_qual_inet}, our prompts are intentionally coarse to align with our primary objective of evaluating catastrophic forgetting and inter-concept interactions, rather than fine-grained image fidelity. However, we acknowledge the need for a better metric. Furthermore, while FLLP (Ours) selectively utilizes optimal parent groups rather than all of the previously learned concepts, the overall parameter count remains unchanged when compared to CIDM~\cite{cidm}. Although this does not hinder performance, we note that no parameter reduction is achieved through this selective mechanism. Finally, our approach requires tuning of the threshold parameter $\beta$ to maximize performance. However, this is the sole hyperparameter involved.


\end{document}